\documentclass[letterpaper,10pt,journal,twoside]{IEEEtran}

\IEEEoverridecommandlockouts %

\pdfoutput=1

\usepackage{amssymb}
\usepackage{amsmath}
\usepackage{amsfonts}       %
\usepackage{booktabs}       %
\usepackage{balance}
\usepackage[utf8]{inputenc} %
\usepackage[T1]{fontenc}    %
\usepackage[pdftex, pdfstartview={FitV}, pdfpagelayout={TwoColumnLeft},bookmarksopen=true,plainpages = false, colorlinks=true, linkcolor=black, citecolor = black, urlcolor = black,filecolor=black , pagebackref=false,hypertexnames=false, plainpages=false, pdfpagelabels ]{hyperref}
\usepackage{url}            %
\usepackage{nicefrac}       %
\usepackage{microtype}      %
\usepackage[font=scriptsize]{caption}
\usepackage{subcaption}
\usepackage{graphicx}
\usepackage{epstopdf} %
\usepackage{mathtools}
\usepackage{marginnote}
\usepackage{colortbl}
\usepackage[dvipsnames]{xcolor}
\usepackage{float}
\usepackage[capitalize]{cleveref}
\usepackage[]{algorithmic}
\usepackage{makecell}
\usepackage{multirow}
\usepackage{paralist}
\floatstyle{boxed} \floatname{algorithm}{Algorithm}
\newfloat{algorithm}{t}{loa}[section]
\usepackage{xcolor}
\usepackage[nolist,nohyperlinks]{acronym}
\usepackage[sort,compress]{cite}

\usepackage{setspace}
\usepackage{tikz}
\usepackage{amsmath}
\usepackage{color}
\usepackage{soul}
\usepackage{xspace}
\usepackage[capitalize]{cleveref}
\usepackage{dsfont}
\usepackage{bm}
\usepackage{bbm}
\usepackage{nicefrac}  %
\usepackage{amsfonts}
\usepackage{amssymb}
\usepackage{array}
\usepackage{mathtools}
\usepackage{caption}
\usepackage{subcaption}
\usepackage{fontawesome}
\usepackage{thmtools}
\usepackage{thm-restate}
\usepackage{lipsum}
\usepackage{pifont}
\usepackage{makecell}
\usepackage{multirow}
\usepackage{afterpage}
\usepackage{tabularx} 
\usepackage{url}
\usepackage[normalem]{ulem}
\floatstyle{boxed} \floatname{algorithm}{Algorithm}
\newfloat{algorithm}{t}{loa}[section]
\usepackage{xcolor}
\usepackage{amsmath}
\usepackage{cleveref}
\usepackage{wrapfig}  %
\usepackage{textcomp}

\newcommand{\NoRed}{\textbf{\textcolor{red}{No}} \cellcolor{Red!10}}
\newcommand{\YesGreen}{\textbf{\textcolor{ForestGreen}{Yes}} \cellcolor{LimeGreen!25}}

\Crefname{figure}{Figure}{Figures}
\crefname{figure}{Figure}{Figures}
\crefname{table}{Table}{Tables}
\crefformat{table}{Table~#1}
\crefformat{equation}{Equation (#1)}
\Crefformat{equation}{Equation (#1)}
\crefformat{algorithm}{Algorithm~#1}
\crefformat{appendix}{Appendix~#1}
\Crefformat{appendix}{Appendix~#1}
\crefformat{appendices}{Appendices~#1}
\Crefformat{appendices}{Appendices~#1}

\newcommand\sdots{\hbox to 1em{.\hss.\hss.}} %
\DeclareMathAlphabet\mathbfcal{OMS}{cmsy}{b}{n}  %

\newcommand{\xdes}{\mathbf{x}^\text{des}}    %
\newcommand{\xsafe}{\bar{\mathbf{x}}}   %
\newcommand{\usafe}{\bar{\mathbf{u}}}   %
\newcommand{\xest}{\hat{\mathbf{x}}}    %
\newcommand{\eest}{\boldsymbol{\xi}^{\text{est}}}  %
\newcommand{\ectrl}{\boldsymbol{\xi}^{\text{ctrl}}}%

\newcommand{\sourcedomain}{\mathcal{S}}
\newcommand{\targetdomain}{\mathcal{T}}

\newif\ifcomments
\commentstrue

\ifcomments
	\newcommand{\aXX}[1]{\color{OliveGreen}AT: (#1)\color{black}\xspace}  %
	\newcommand{\XX}[1]{\color{red}JH: (#1)\color{black}\xspace}  %
\else
    \newcommand{\dXX}[1]{}  %
	\newcommand{\aXX}[1]{}  %
	\newcommand{\XX}[1]{}  %
\fi

\newcommand{\PreserveBackslash}[1]{\let\temp=\\#1\let\\=\temp}
\newcolumntype{C}[1]{>{\PreserveBackslash\centering}p{#1}}
\newcolumntype{R}[1]{>{\PreserveBackslash\raggedleft}p{#1}}
\newcolumntype{L}[1]{>{\PreserveBackslash\raggedright}p{#1}}

\crefformat{equation}{(#2#1#3)}

\renewcommand{\b}[1]{{#1}}

\title{\LARGE \bf
Tube-NeRF: Efficient Imitation Learning of Visuomotor Policies from MPC using Tube-Guided Data Augmentation and NeRFs\\ 
}

\author{Andrea Tagliabue, Jonathan P. How%
\thanks{Authors with the Laboratory for Information and Decision Systems (LIDS), MIT. \tt\{atagliab, jhow\}@mit.edu}%
\vspace*{-3em}
\thanks{Research supported by the AFOSR MURI FA9550-19-1-0386.}
}%

\begin{document}
\bstctlcite{IEEEexample:BSTcontrol}
\maketitle
\thispagestyle{empty}
\pagestyle{empty}

\begin{abstract}
Imitation learning (IL) can train computationally-efficient sensorimotor policies from a resource-intensive Model Predictive Controller (MPC), but it often requires many samples, leading to long training times or limited robustness.  
To address these issues, we combine IL with a variant of robust MPC that accounts for \textit{process and sensing} uncertainties, and we design a data augmentation (DA) strategy that enables \textit{efficient} learning of \textit{vision-based} policies. 
The proposed DA method, named Tube-NeRF, leverages Neural Radiance Fields (NeRFs) to generate novel synthetic images, and uses properties of the robust MPC (the \textit{tube}) to select relevant views and to efficiently compute the corresponding actions. 
We tailor our approach to the task of localization and trajectory tracking on a multirotor, by learning a \textit{visuomotor} policy that generates control actions using images from the onboard camera as only source of horizontal position.  
Numerical evaluations show $80$-fold increase in demonstration efficiency and a $50\%$ reduction in training time over current IL methods. Additionally, our policies successfully transfer to a real multirotor, achieving low tracking errors despite large disturbances, with an onboard inference time of only $1.5$ ms. Video: \href{https://youtu.be/_W5z33ZK1m4}{\texttt{\color{blue}https://youtu.be/\_W5z33ZK1m4}}

\end{abstract}

\section{INTRODUCTION}
\ac{IL} \cite{pomerleau1989alvinn, argall2009survey, ross2011reduction} has been extensively employed to train \textit{sensorimotor} \ac{NN} policies for computationally-efficient onboard sensing, planning and control on mobile robots. Training data is \b{commonly provided by a computationally expensive \ac{MPC} \cite{loquercio2019deep, kaufmann2020deep, pan2020imitation}, acting as \textit{expert} demonstrators. The resulting policies produce commands from raw sensory inputs, simultaneously bypassing the computational cost of solving large optimization problems in \ac{MPC}, as well as sensing and  localization.}

However, one of the fundamental limitations of existing \ac{IL} methods employed to train sensorimotor policies (\ac{BC}) \cite{argall2009survey}, DAgger \cite{ross2011reduction}) is the overall number of demonstrations that must be collected \b{from \ac{MPC}}. 
\b{This inefficiency is rooted in training/deployment data distribution mismatches (\textit{covariate shift} \cite{ross2011reduction})
and is worsened by uncertainties at deployment \cite{laskey2017dart, tagliabue2021demonstration}.
Although strategies that match the uncertainties between the training and deployment domains (\ac{DR} \cite{peng2018sim, tobin2017domain, loquercio2019deep, kaufmann2020deep}) improve robustness, they create challenges in data- and computational-efficiency by repeatedly querying a computationally-expensive expert under different disturbances.}

\begin{figure}[t]
    \centering
    \begin{tikzpicture}
        \node[inner sep=0pt] (figure) at (0,0) {
        \includegraphics[width=0.99\columnwidth]{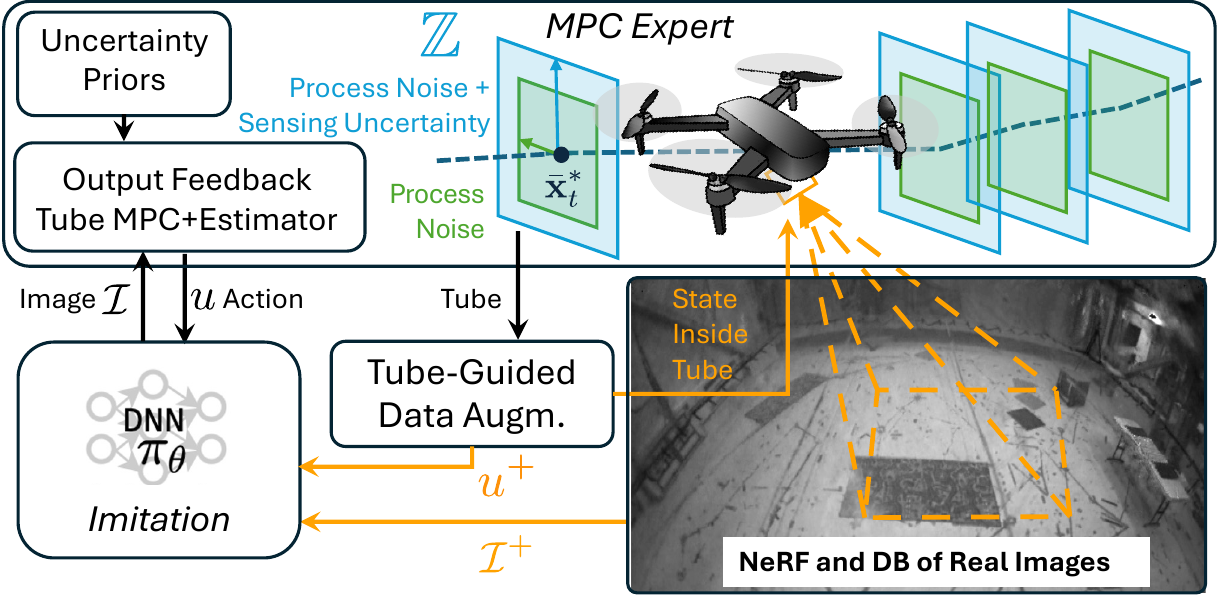}
        };
        \end{tikzpicture}
    
    \caption{
    \b{Tube-NeRF collects a real-world demonstration using output-feedback tube MPC, a robust MPC which accounts for process and sensing uncertainties through its tube cross-section $\mathbb{Z}$. Then, it generates a \acf{NeRF} of the environment from the collected images $\mathcal{I}_t$, and uses the tube's cross-section to guide the selection of synthetic views $\mathcal{I}_t^+$ from the NeRF for data augmentation, while corresponding actions are obtained via the \textit{ancillary controller}, an integral component of the tube MPC framework.}}
    \label{fig:approach_overview}
    \vskip-3ex
\end{figure}

\b{Leveraging high-fidelity simulators on powerful computers in combination with DR avoids demanding data-collection on the real robot}, but it introduces \textit{sim-to-real} gaps that are especially challenging when learning \textit{visuomotor} policies, i.e., that directly use raw pixels as input. For this reason, sensorimotor policies trained in simulation often leverage as input easy-to-transfer visual abstractions, such as feature tracks \cite{kaufmann2020deep}, depth maps \cite{loquercio2021learning, kahn2017plato}, intermediate layers of a \ac{CNN} \cite{9560916}, or a learned latent space \cite{bonatti2020learning}. 
However, all these abstractions discard information that may instead benefit task performance. 

\ac{DA} approaches that augment demonstrations with extra images and stabilizing actions improve robustness and sample efficiency of \ac{IL} \cite{bojarski2016end, giusti2015machine, sharma2022augmenting, muller2018teaching, zhou2023nerf}. However, they
\begin{inparaenum}[(i)]
\item rely on handcrafted heuristics for the selection of extra images and the generation of corresponding actions, 
\item do not explicitly account for the effects of uncertainties when generating the extra data \cite{bojarski2016end, giusti2015machine, sharma2022augmenting, muller2018teaching, zhou2023nerf}, and
\item often leverage ad-hoc image acquisition setups for \ac{DA} \cite{giusti2015machine, sharma2022augmenting}.
\end{inparaenum} As a consequence, their real-world deployment has been mainly focused on tasks in 2D \b{(steering a Dubins car \cite{bojarski2016end, sharma2022augmenting})}.

\b{This work, as shown in \cref{fig:approach_overview}, introduces Tube-\acf{NeRF}, a novel \ac{DA} framework for efficient, robust \textit{visuomotor} policy learning from \ac{MPC} that overcomes the aforementioned limitations in \ac{DA}. Building on our prior \ac{DA} strategy \cite{tagliabue2021demonstration} that uses  \ac{RTMPC} for \textit{efficiently} and \textit{systematically} generating additional training data for motor control policies (actions from full state), Tube-\ac{NeRF} enables learning of policies that directly use vision as input, relaxing the constraining assumption in \cite{tagliabue2021demonstration} that full-state information is available at deployment. }

Our new strategy first collects robust task-demonstrations that account for the effects of process \textit{and sensing} uncertainties via an output-feedback variant of \ac{RTMPC} \cite{mayne2006robust}.
Then, \b{it employs} a photorealistic representation of the environment, based on a \ac{NeRF}, to generate \b{synthetic} novel views for \ac{DA}, using the tube of the controller to guide the selection of the extra novel views and sensorial inputs, and using the ancillary controller to efficiently compute the corresponding actions. 
We further enhance \ac{DA} by organizing collected real-world observations in a database and employing the tube to generate queries. This enables usage of real-world data for \ac{DA}, reducing the sim-to-real gap when only synthetic images are used. 
Lastly, we adapt our approach to a multirotor, training a \textit{visuomotor} policy for robust trajectory tracking and localization using onboard camera images and additional measurements of altitude, orientation, and velocity. The generated policy relies solely on images to obtain information on the robot's horizontal position, which is a challenging task due to (1) its high speed (up to $3.5$ m/s), (2) varying altitude, (3) aggressive roll/pitch changes, (4) the sparsity of visual features in our flight space, and (5) the presence of a safety net that moves due to the down-wash of the propellers and that produces semi-transparent visual features above the ground.

\b{
\noindent \textbf{Contributions}: This Letter is an evolution of our previous conference paper \cite{tagliabue2022output}, where 
we demonstrated the capabilities of an output-feedback RTMPC-guided DA strategy in simulation. Now, we present a set of algorithmic changes that enabled the first real-world, real-time deployment of the approach. Specifically, different from \cite{tagliabue2022output}, our new algorithm:
        \begin{itemize}
        \item Utilizes a \ac{NeRF} for photo-realistic novel view synthesis for \ac{DA}, replacing a 3D mesh, and presents a calibration method to integrate the \ac{NeRF} into the DA framework.
        \item Further reduces the \textit{sim-to-real gap} that, although small, still exists in the synthetic images from the NeRF by modifying the \ac{DA} strategy to additionally sample available real-world images inside the tube.
        \item Accounts for visual changes in the environment by introducing in training randomizations in image space.
        \item Includes noisy altitude measurements in the policy's input.
        \item Accounts for camera calibration errors by introducing perturbations in the camera extrinsic during DA. 
        \item Employs a more capable but computationally efficient visuomotor \ac{NN} architecture for onboard deployment.
    \end{itemize}
Therefore, this work contributes procedures to: 
\begin{itemize}
\item Efficiently (demonstrations, training time) learn of a \textit{sensorimotor} policy from \ac{MPC} using a DA strategy grounded in the output feedback \ac{RTMPC} framework theory, unlike previous \ac{DA} methods that rely on handcrafted heuristics.
\item Apply our method for tracking and localization on a multirotor via images, altitude, attitude and velocity data.
\item Achieve the first-ever real-time deployment of our approach, demonstrating (in more than $30$ flights) successful agile trajectory tracking with policies learned from a \textbf{single demonstration} that use onboard fisheye images to infer the horizontal position of a multirotor despite aggressive 3D motion, and subject to a variety of sensing and dynamics disturbances. Our policy has an average inference time of only $1.5$ ms onboard a small GPU (Nvidia Jetson TX2) and is deployed at $200$ Hz. The policy is trained using a single demonstration.
\end{itemize}
}

\vskip-10ex

\section{RELATED WORKS}
\noindent
\textbf{Robust/efficient \ac{IL} of sensorimotor policies.} \cref{tab:state_of_the_art_comparison} presents state-of-the-art approaches for sensorimotor policy learning from demonstrations (from \ac{MPC} or humans), focusing  on mobile robots. DA-NeRF is the only method that
\begin{inparaenum}[(i)]
    \item explicitly accounts for uncertainties, 
    \item is efficient to train, and 
    \item does not require visual abstractions  
    \item nor specialized data collection setups.
\end{inparaenum}
Related to our research, \cite{zhou2023nerf} employs a \ac{NeRF} for \ac{DA} from human demonstrations for manipulation, but uses heuristics to select relevant views, without explicitly accounting for uncertainties. Tube-NeRF uses properties of robust MPC to select relevant views and actions for \ac{DA}, accounting for uncertainties. In addition, we incorporate available real-world images for \ac{DA}, further reducing the sim-to-real gap. 

\begin{table}[!t]
    \vskip2ex
    \renewcommand{\arraystretch}{1.4}
    \scriptsize
    \caption{Approaches that learn visuomotor policies from demonstrations. 
    We highlight that (i) training the policy entirely in simulation often requires visual-abstractions for real-world deployments (e.g., \cite{kaufmann2020deep}), losing however important information about the environment. Instead, approaches that directly use images (ii) benefit from data collection in the real-world, but require a large number of demonstrations (e.g., \cite{pan2020imitation}), or (iii) leverage \ac{DA} strategies that however use specialized data collection equipment and their deployment has been focused on ground robots (2D domain) (~\cite{bojarski2016end, sharma2022augmenting}) or to generate low-dimensional, discrete actions for aerial tasks (e.g., move left-right \cite{giusti2015machine}). Additionally, all the considered \ac{DA} approaches employ ad-hoc heuristics to select extra sensorial data and/or to compute the corresponding actions.}
    \vskip-1ex
    \begin{centering}
    \label{tab:state_of_the_art_comparison}
    \resizebox{1.0\columnwidth}{!}{
    \begin{tabular}{
    >{\centering\arraybackslash}m{0.23\columnwidth} 
    >{\centering\arraybackslash}m{0.1\columnwidth}
    >{\centering\arraybackslash}m{0.1\columnwidth}
    >{\centering\arraybackslash}m{0.14\columnwidth}
    >{\centering\arraybackslash}m{0.07\columnwidth}
    >{\centering\arraybackslash}m{0.12\columnwidth}
    >{\centering\arraybackslash}m{0.18\columnwidth} %
    >{\centering\arraybackslash}m{0.2\columnwidth} %
    }
    \toprule 
    \textbf{Method} & 
    \textbf{Domain of training data} &
    \textbf{Policy directly uses images} & 
    \textbf{No special data collection equipment} & 
    \textbf{Demo.-efficient} & 
    \textbf{Explicitly accounts for uncertainties} & 
    \textbf{Avoids Hand-Crafted Heuristics for Data Augm.} & %
    \textbf{Real-world deployment (2D/3D, Domain)} %
    \tabularnewline
    \midrule
    \cite{zhang2016learning} (\textbf{MPC-GPS}) & Sim. & \YesGreen{}{} & \YesGreen{}{} & \NoRed{} & \NoRed{} & N.A. & \NoRed{} (3D, Aerial) \tabularnewline
    \cline{0-0}
    \cite{kahn2017plato} (\textbf{PLATO}) & Sim. & \NoRed{} & \YesGreen{} & \YesGreen{} & \NoRed{} & N.A. & \NoRed{} (3D, Aerial) \tabularnewline
    \cline{0-0}
    \cite{muller2018teaching} \textbf{(BC+DA)} & Sim. & \YesGreen{} & \YesGreen{} & \YesGreen{} & \NoRed{} & \NoRed{} & \NoRed{} (3D, Aerial) \tabularnewline
    \cline{0-0} 
    \cite{kaufmann2020deep} (\textbf{DAgger+DR}) & Sim & \NoRed{} & \YesGreen{} & \NoRed {} & \YesGreen{} & N.A. & \YesGreen{} (3D, Aerial) \cellcolor{LimeGreen!25} \tabularnewline
    \cline{0-0}
    \cite{pan2020imitation} (\textbf{DAgger}) & Real & \YesGreen{} & \YesGreen{} & \NoRed{} & \NoRed{} & N.A. & Yes (2D, Ground) \tabularnewline
    \cline{0-0}
    \cite{sharma2022augmenting} (\textbf{DAgger+DA}) & Real & \YesGreen{} & \NoRed{} & \YesGreen{} & \NoRed{} & \NoRed{} & Yes (2D, Ground) \tabularnewline
    \cline{0-0}
    \cite{bojarski2016end} (\textbf{BC+DA}) & Real & \YesGreen{} & \NoRed{} & \YesGreen{} & \NoRed{} & \NoRed{} & Yes (2D, Ground) \tabularnewline
    \cline{0-0}
    \cite{giusti2015machine} (\textbf{BC+DA}) & Real & \YesGreen{} & \NoRed{} & \YesGreen{} & \NoRed{} & \NoRed{} & \YesGreen{} (3D, Aerial) \tabularnewline
    \cline{0-0}
    \cite{zhou2023nerf} (\textbf{SPARTAN}) & Real & \YesGreen{} & \YesGreen{} & \YesGreen{} & \NoRed{} & \NoRed{} & Yes (3D, Arm) \tabularnewline
    \hline
    \textbf{Tube-NeRF} (proposed) & Real & \YesGreen{} & \YesGreen{} &  \YesGreen{} & \YesGreen{} & \YesGreen{} & \YesGreen{} (3D, Aerial) \cellcolor{LimeGreen!25} \tabularnewline
    \bottomrule
    \end{tabular}
    }
    \par
    \end{centering}
    \vskip-5ex
\end{table}

\noindent
\textbf{Novel view synthesis with \acp{NeRF}.}
\acp{NeRF} \cite{mildenhall2021nerf} enable efficient \cite{muller2022instant} and photorealistic novel view synthesis by directly optimizing the photometric accuracy of the reconstructed images, in contrast to traditional 3D photogrammetry methods (e.g., for 3D meshes). This provides accurate handling of transparency, reflective materials, and lighting conditions.
Ref.~\cite{byravan2023nerf2real} employs a \ac{NeRF} to create a simulator for learning legged robot control policies from RGB images using \ac{RL}, \b{using a specialized camera for data collection, unlike our work}. %
Ref.~\cite{adamkiewicz2022vision} uses a \ac{NeRF} for estimation, planning, and control on a drone by querying the \ac{NeRF} online, but this results in $1000\times$ higher computation time\footnote{Control and estimation in \cite{adamkiewicz2022vision} require $6.0$s on a NVIDIA RTX3090 GPU ($10,496$ CUDA cores, $24$ GB VRAM), while ours requires $1.5$ms on a much smaller Jetson TX2 GPU ($256$ CUDA cores, $8$ GB shared RAM).} than our policy.

\noindent
\textbf{Output feedback RTMPC.} \ac{MPC} \cite{borrelli2017predictive} solves a constrained optimization problem that uses a model of the system dynamics to plan for actions that satisfy state and actuation constraints. 
\ac{RTMPC} assumes that the system is subject to additive, bounded \textit{process} uncertainty (disturbances, model errors) and employs an auxiliary (ancillary) controller that maintains the system within some known distance (cross-section of a tube) of the plan \cite{mayne2006robust}. 
\textit{Output feedback} \ac{RTMPC} \cite{mayne2006robust, lorenzetti2020simple} in addition accounts for the effects of \textit{sensing} uncertainty (noise, estimation errors) by increasing the cross-section of the tube. Our method uses an output-feedback RTMPC for data collection but bypasses its computational cost at deployment by learning a \ac{NN} policy.

\section{PROBLEM FORMULATION}
Our goal is to \textit{efficiently} train a \ac{NN} visuomotor policy $\pi_\theta$ (\textit{student}), with parameters $\theta$, that tracks a desired trajectory on a mobile robot \b{(multirotor)}. $\pi_\theta$ takes as input images, which are needed to extract partial state information \b{(horizontal position, in our evaluation)}, and other measurements. 
The trained policy, denoted $\pi_{\hat{\theta}^*}$, needs to be robust to uncertainties encountered in the deployment domain $\targetdomain$. It is trained using demonstrations from a model-based controller (\textit{expert}) collected in a source domain $\sourcedomain$ that presents only a subset of the uncertainties in $\targetdomain$.

\noindent
\textbf{Student policy.}
The student policy has the form:
\begin{equation}
\label{eq:policy} \small
    \mathbf u_t = \pi_\theta(\mathbf{o}_t, \mathbf{X}^\text{des}_t),
\end{equation}
and it generates deterministic, continuous actions $\mathbf u_t$ to track a desired $N+1$ steps ($N > 0$) trajectory $\mathbf{X}^\text{des}_t = \{\xdes_{0|t},\dots,\xdes_{N|t}\}$.
$\mathbf{o}_t = (\mathcal{I}_t, \mathbf{o}_{\text{other},t})$ are noisy sensor measurements comprised of an image $\mathcal{I}_t$ from an onboard camera, and other measurements $\mathbf{o}_\text{other,t}$ (altitude, attitude, velocity, \b{in our evaluation}). %

\noindent
\textbf{Expert.} \textit{Process model:}
\b{The considered robot dynamics are}: 
\begin{equation} \small
    \mathbf x_{t+1} = \mathbf A \mathbf x_{t} + \mathbf B  \mathbf u_{t} + \mathbf w_{t} \\
\label{eq:linearized_dynamics}
\end{equation}
where $\mathbf x_t  \in \mathbb{X} \subset \mathfrak{R}^{n_x}$ is the state, and $\mathbf u_t \in \mathbb{U} \subset \mathfrak{R}^{n_u}$ the control inputs. The robot is subject to state and input constraints $\mathbb{X}$ and $\mathbb{U}$, assumed convex polytopes containing the origin \cite{mayne2006robust}. 
$\mathbf w_t \in \mathbb{W}_{\targetdomain} \subset \mathfrak{R}^{n_x}$ in \eqref{eq:linearized_dynamics} captures time-varying additive \textit{process} uncertainties in $\targetdomain$, such as
\begin{inparaenum}[(i)]
\item disturbances \b{(wind/payloads for a UAV)}
\item and model changes/errors \b{(linearization errors and poorly known parameters)}.
\end{inparaenum} 
$\mathbf w_t$ is unknown, but the polytopic bounded set $\mathbb{W}_{\targetdomain}$ is assumed known\cite{mayne2006robust}. 
\textit{Sensing model:} The expert has access to (i) the measurements $\mathbf{o}_{\text{other},t}$, and (ii) a vision-based position estimator $g_\text{cam}$ that outputs noisy measurements $\mathbf o_{\text{pos.xy}} \in \mathfrak{R}^2$ of the horizontal position $\boldsymbol{p}_{{xy},t} \in \mathfrak{R}^2$ of the robot:
\begin{equation}
\label{eq:position_estimator}
    \mathbf o_{\text{pos,xy}, t} =  g_{\text{cam}}(\mathcal{I}_t)  = \boldsymbol{p}_{{xy},t} + \mathbf v_{\text{\text{cam}},t}, 
\end{equation} where $\mathbf{v}_{\text{cam},t}$ is the associated sensing uncertainty. 
The measurements available to the expert are denoted $\bar{\mathbf{o}}_t \in \mathfrak{R}^{n_o}$, and they map to the robot state via:
\begin{equation}
\label{eq:obs_model} \small
    \bar{\mathbf{o}}_t = 
    \begin{bmatrix}
        \mathbf o_{\text{pos.xy}, t}^T, \mathbf o_{\text{other}, t}^T
    \end{bmatrix}^T
    = \mathbf C
    \mathbf x_t 
    + \mathbf v_t,
\end{equation}
where $\mathbf C \in \mathfrak{R} ^{n_o \times n_x}$. $\mathbf v_t = [\mathbf v_{\text{\text{cam}},t}^T, \mathbf v_{\text{other},t}^T]^T \in \mathbb{V}_\targetdomain \subset \mathfrak{R}^{n_o}$ is additive \textit{sensing} uncertainty (e.g., noise, biases) in $\targetdomain$. $\mathbb{V}_\targetdomain$ is a known bounded set obtained via system identification, and/or via prior knowledge on the accuracy of $g_\text{cam}$.

\noindent
\textit{State estimator.} We assume the expert uses a state estimator: 
\label{sec:state_observer}
\begin{equation} \label{eq:estimator} \small
    \xest_{t+1} = \mathbf A \xest_t + \mathbf B \mathbf u_t + \mathbf L ( \bar{\mathbf{o}}_t - \hat{\mathbf o}_t), \quad \hat{\mathbf o}_t = \mathbf C \xest_t,
\end{equation}
where 
$\xest_t \in \mathfrak{R}^{n_x}$ is the estimated state, and $\mathbf L \in \mathfrak  R^{n_x \times n_o}$ is the observer gain, set so that $\mathbf A - \mathbf L \mathbf C$ is Schur stable. The observability index of the system $(\mathbf A, \mathbf C)$ is assumed to be $1$, meaning that full state information can be retrieved from a single noisy measurement. In this case, the observer plays the critical role of filtering out the effects of noise and sensing uncertainties. Additionally, we assume that the state estimation dynamics and noise sensitivity of the learned policy will approximately match the ones of the observer.

\section{METHODOLOGY}
\textbf{Overview.} Tube-NeRF collects trajectory tracking demonstrations in the source domain $\sourcedomain$ using an output feedback \ac{RTMPC} expert combined with a state estimator \cref{eq:estimator}, and \ac{IL} methods (DAgger or \ac{BC}). The chosen output feedback \ac{RTMPC} framework is based on \cite{mayne2006robust, lorenzetti2020simple}, with its objective function modified to track a trajectory (\cref{sec:tube_mpc}), and is designed according to the priors on process \textit{and} sensing uncertainties at deployment ($\targetdomain$). 
Then, Tube-NeRF uses properties of the expert to design an efficient \ac{DA} strategy, the key to overcoming efficiency and robustness challenges in \ac{IL} (\cref{sec:data_augmentation}). The framework is then tailored to a multirotor leveraging a \ac{NeRF} as part of the proposed \ac{DA} strategy (\cref{sec:application_to_vision_flights}).

\subsection{Output feedback robust tube MPC for trajectory tracking} \label{sec:tube_mpc}
Output feedback \ac{RTMPC} for trajectory tracking regulates the system in \cref{eq:linearized_dynamics} and \cref{eq:estimator} along a given \textit{reference} trajectory $\mathbf{X}^\text{des}_t$, while satisfying state and actuation constraints $\mathbb X, \mathbb U$ regardless of sensing uncertainties ($\mathbf v$, \cref{eq:obs_model}) and process noise ($\mathbf w$, \cref{eq:linearized_dynamics}). 

\noindent
\textbf{Preliminary (set operations):} Given the convex polytopes $\mathbb{A} \subset \mathfrak{R}^{n}, \mathbb{B} \subset \mathfrak{R}^{n}$ and $\mathbf{M} \in \mathfrak{R}^{m \times n}$, we define:
\begin{itemize}
\item Minkowski sum: $\mathbb{A} \oplus \mathbb{B} \coloneqq \{\mathbf a + \mathbf b \in \mathfrak{R}^n \:|\: \mathbf a \in \mathbb{A}, \: \mathbf b \in \mathbb{B} \}$
\item Pontryagin diff.: $\mathbb{A} \ominus \mathbb{B} \coloneqq \{\mathbf c \in \mathfrak{R}^n \:|\: \mathbf{c + b} \in \mathbb{A}, \forall \mathbf  b \in \mathbb{B} \}$ 
\item Linear mapping: $\mathbf M \mathbb{A} \coloneqq \{\mathbf{M} \mathbf a \in \mathfrak{R}^m \:|\: \mathbf a \in \mathbb{A}\}$.
\end{itemize}

\noindent
\textbf{Optimization problem.} At every timestep $t$, \ac{RTMPC} solves:
\begin{equation} \label{eq:ofrtmpc}
\small
\begin{split}
    \mathbf{\bar{U}}_t^*, \mathbf{\bar{X}}_t^*
    = \underset{\mathbf{\bar{U}}_t, \mathbf{\bar{X}}_t}{\text{argmin}} & 
        \| \mathbf e_{N|t} \|^2_\mathbf{P} + 
        \sum_{i=0}^{N-1} 
            \| \mathbf e_{i|t} \|^2_\mathbf{Q} + 
            \| \mathbf u_{i|t} \|^2_\mathbf{R} \\
    \text{subject to} \:\: & \xsafe_{i+1|t} = \mathbf A \xsafe_{i|t} + \mathbf B \usafe_{i|t}, \:\: i = 0, \dots, N-1   \\
    & \xsafe_{i|t} \in \bar{\mathbb{X}}, \:\: \usafe_{i|t} \in \bar{\mathbb{U}}, \\
    & \mathbf{\bar{x}}_{N|t} \in \bar{\mathbb{X}}_N, \:\: 
    \xest_t \in \xsafe_{0|t} \oplus 
            \begin{bmatrix}
            \boldsymbol{0}_{n_x},\!\!\!\!&\!\!\! \boldsymbol{I}_{n_x} \\
            \end{bmatrix}\mathbb{S},\\
\end{split}
\end{equation}
where $\mathbf e_{i|t} = \xsafe_{i|t} - \xdes_{i|t}$ represents the trajectory tracking error, $\mathbf{\bar{X}}_t = \{\xsafe_{0|t},\dots,\xsafe_{N|t}\}$ and $\mathbf{\bar{U}}_t = \{\usafe_{0|t},\dots,\usafe_{N-1|t}\}$ are \textit{safe} reference state and action trajectories, and $N+1$ is the length of the planning horizon. The positive semi-definite matrices $\mathbf{Q}$ (size $n_x \times n_x$) and $\mathbf{R}$ (size $n_u \times n_u$) are tuning parameters, $\| \mathbf e_{N|t} \|^2_\mathbf{P}$ is a terminal cost (obtained by solving the infinite-horizon optimal control problem with $\mathbf A$, $\mathbf B$, $\mathbf Q$, $\mathbf R$) and $\mathbf{\bar{x}}_{N|t} \in \bar{\mathbb{X}}_N$ is a terminal state constraint. 

\noindent
\textbf{Ancillary controller.} The control input $\mathbf{u}_t$ is obtained via:
\begin{equation}
\small
\label{eq:ancillary_controller}
    \mathbf u_t = \mathbf \usafe^*_{t} + \mathbf K (\xest_t - \xsafe^*_{t}),
\end{equation}
where $\usafe^*_t \coloneqq \usafe^*_{0|t}$ and $\xsafe^*_t \coloneqq \xsafe^*_{0|t}$, and $\mathbf{K}$ is computed by solving the LQR problem with $\mathbf A$, $\mathbf B$, $\mathbf Q$, $\mathbf{R}$. 
This controller maintains the system inside a set $\mathbb{Z} \oplus \xsafe_t^*$ (``cross-section'' of a \textit{tube} centered around $\xsafe_t^*$), regardless of the uncertainties. %

\noindent
\textbf{Tube and robust constraints.}
Process and sensing uncertainties are taken into account by tightening the constraints $\mathbb{X}$, $\mathbb{U}$, obtaining $\bar{\mathbb{X}}$ and $\bar{\mathbb{U}}$ in \cref{eq:ofrtmpc}. The amount by which $\mathbb{X}$, $\mathbb{U}$ are tightened depends on the cross-section of the tube $\mathbb{Z}$, which is computed \b{(see \cite{mayne2006robust}) by considering the closed-loop system formed by the ancillary controller \cref{eq:ancillary_controller}, the nominal dynamics in \cref{eq:linearized_dynamics} and the observer \cref{eq:estimator}}. \textit{Sensing uncertainties} in $\mathbb{V}_\targetdomain$ and \textit{process noise} in $\mathbb{W}_\targetdomain$ introduce two sources of errors in such system: 
\begin{inparaenum}[a)]
\item the state estimation error $\eest_t \coloneqq \mathbf x_t - \xest_t$, and
\item the control error $\ectrl_t \coloneqq \xest_t - \xsafe_t^*$.
\end{inparaenum}
These errors can be combined in a vector $\boldsymbol{\xi}_t = [{\eest_t}^T, {\ectrl_t}^T ]^T$ whose dynamics evolve according to (see \cite{lorenzetti2020simple}):
\begin{equation} 
\small
\label{eq:error}
\boldsymbol{\xi}_{t+1}  = \mathbf A_\xi \boldsymbol{\xi}_t  + \boldsymbol{\delta}_t, \; \boldsymbol{\delta}_t \in \mathbb{D}\\
\end{equation}
\begin{equation*}
\small
\mathbf A_\xi = \begin{bmatrix}
\mathbf A- \mathbf L \mathbf C & \mathbf{0}_{n_x} \\ \mathbf L \mathbf C & \mathbf A + \mathbf B \mathbf K
\end{bmatrix}, \quad 
\mathbb{D} = \begin{bmatrix}
\mathbf I_{n_x} & -\mathbf L \\ \mathbf{0}_{n_x} & \mathbf L 
\end{bmatrix}\begin{bmatrix}
\mathbb{W}_\targetdomain \\ \mathbb{V}_\targetdomain
\end{bmatrix}.
\label{eq:autonomous_system}
\end{equation*}

\begin{figure}
\centering
        \begin{tikzpicture}
        \node[inner sep=0pt] (figure) at (0,0) {\includegraphics[trim={95, 40, 75, 175}, clip, width=\linewidth]{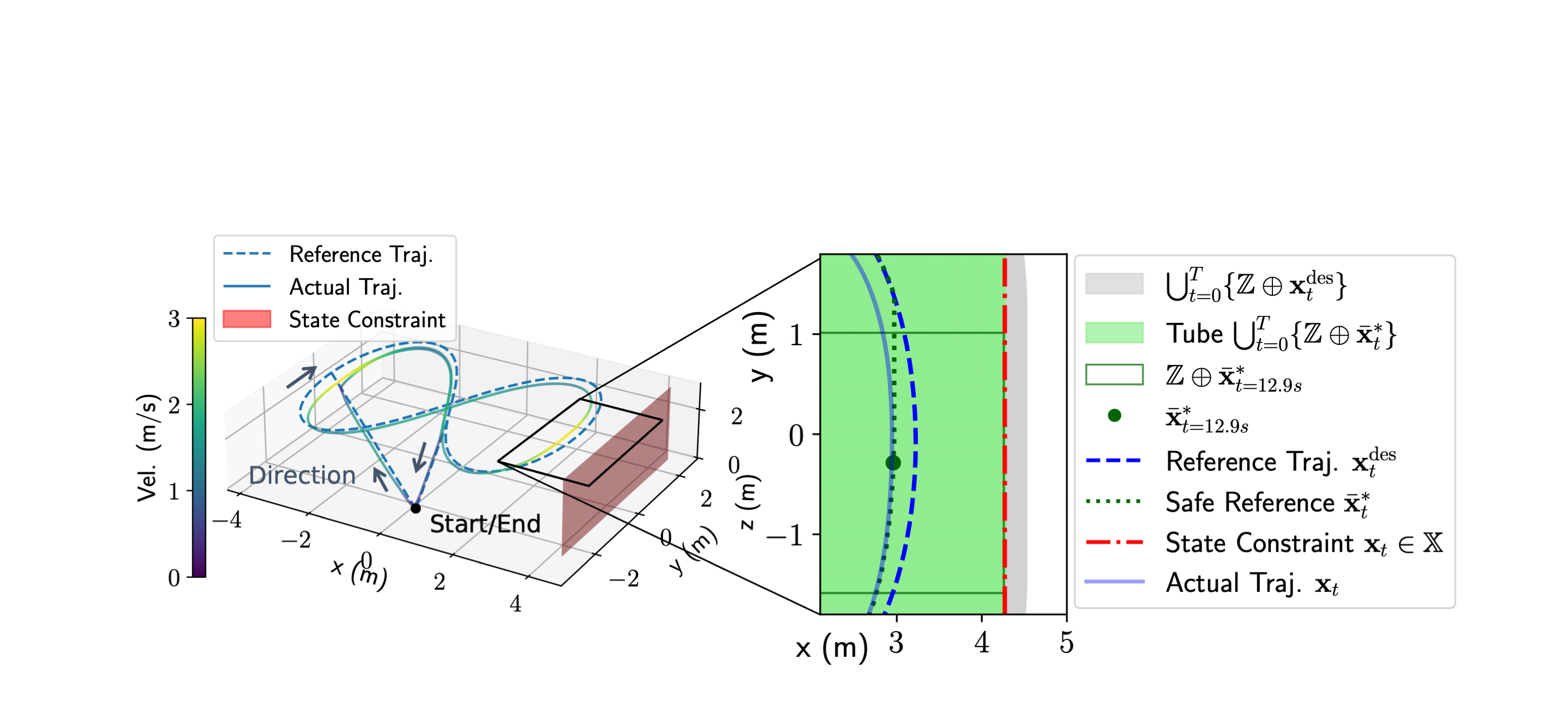}};
        \end{tikzpicture}
        
        \caption{Output feedback RTMPC generates and tracks a safe reference to satisfy constraints.}
    \label{fig:ofrtmpc_example}
    \vskip-3ex
\end{figure}

By design, $\mathbf A_\xi$ is a Schur-stable dynamic system, and it is subject to uncertainties from the convex polytope $\mathbb{D}$. Then, it is possible to compute the \textit{minimal} Robust Positive Invariant (RPI) set $\mathbb{S}$, that is the \textit{smallest} set satisfying:
\begin{equation}
\small
    \boldsymbol{\xi}_0 \in \mathbb{S} \implies \boldsymbol{\xi}_t \in \mathbb{S}, \; \forall \; \boldsymbol \delta_t \in \mathbb{D}, \; t > 0.
\end{equation} 
$\mathbb{S}$ represents the possible set of state estimation and control errors caused by uncertainties and is used to compute $\bar{\mathbb{X}}$ and $\bar{\mathbb{U}}$. Specifically, the error between the true state $\mathbf x_t$ and the reference state $\xsafe_t^*$ is:
$
\boldsymbol{\xi}^{\text{tot}} \coloneqq \mathbf{x}_t - \xsafe_t^* = \boldsymbol{\xi}^{\text{ctrl}} + \boldsymbol{\xi}^{\text{est}}.
$
As a consequence, the effects of noise and uncertainties can be taken into account by tightening the constraints of an amount:
\begin{equation}
\label{eq:tube} \small
\bar{\mathbb{X}} \coloneqq \mathbb{X}\ominus
\mathbb{Z}, \;
\mathbb{Z} = 
\begin{bmatrix}
\boldsymbol{I}_{n_x} & \boldsymbol{I}_{n_x} \\
\end{bmatrix}\mathbb{S},
\quad \bar{\mathbb{U}} \coloneqq \mathbb{U} \ominus\begin{bmatrix}
\boldsymbol{0}_{n_x} & \mathbf K
\end{bmatrix} \mathbb{S}.
\end{equation}
$\mathbb{Z}$ (cross-section of a \textit{tube}) is the set of possible deviations of the true state $\mathbf{x}_t$ from the safe reference $\xsafe_t^*$.
\b{\textbf{Computing $\mathbb{S}$.} While accurately computing the minimal RPI set $\mathbb{S}$ for high-dimensional systems can be challenging \cite{lorenzetti2020simple}, for simplicity, we efficiently obtain $\mathbb{S}$ from $\mathbb{W}_\targetdomain$ and $\mathbb{V}_\targetdomain$ via Monte Carlo simulations of \cref{eq:error}, uniformly sampling instances of the uncertainties, and by computing an outer axis-aligned bounding box of the trajectories of $\boldsymbol{\xi}$. In addition, we treat linearization errors and future changes in the reference trajectory as an additional source of uncertainty, computing the tube based on an increased  process uncertainty prior $\bar{\mathbb{W}}_\targetdomain$, and numerically validating that the resulting expert is robust. While this procedure is approximate, it was found computationally tractable and useful at estimating tubes with an adequate level of conservativeness.} \cref{fig:ofrtmpc_example} shows an example of this controller for trajectory tracking on a multirotor, highlighting changes to the reference trajectory to respect state constraint under uncertainties.  

\subsection{Tube-guided \ac{DA} for visuomotor learning}
\label{sec:data_augmentation}
\noindent
\textbf{IL objective.} 
We denote the \textit{expert} output feedback RTMPC in \cref{eq:ofrtmpc}, \cref{eq:ancillary_controller} and the state observer in \cref{eq:estimator} as $\pi_{\theta^*}$. %
The goal is to design an \ac{IL} and \ac{DA} strategy to efficiently learn the parameters $\hat{\theta}^*$ of the policy \cref{eq:policy}, collecting demonstrations from the expert $\pi_{\theta^*}$. 
In \ac{IL}, this objective consists in minimizing the MSE loss:
\[
\small
    \hat{\theta}^* \! \in \! \text{arg} \! \min_{\theta}\! \mathbb{E}_{p(\tau|\pi_\theta, \targetdomain)}\!
\left[\frac{1}{T}\sum_{t=0}^{T-1}\| \pi_\theta(\mathbf o_t, \mathbf{X}_t^\text{des})\!\!-\!\!\pi_{\theta^*}(\mathbf o_t, \mathbf{X}_t^\text{des})\|_2^2
\right]
\]%
where $\tau:= \{(\mathbf o_t, \mathbf u_t, \mathbf X^\text{des}_t)\! \mid \!\!t\!=\!0,\!\dots,\! T\}$ is a $T+1$ step (observation, action, reference) trajectory sampled from the distribution $p(\tau|\pi_\theta, \targetdomain)$. Such a distribution represents all the possible trajectories induced by the \textit{student} policy $\pi_{\theta}$ in the deployment environment $\targetdomain$. %
As observed in \cite{laskey2017dart, tagliabue2021demonstration}, the presence of uncertainties in $\targetdomain$ makes \ac{IL} challenging, as demonstrations are usually collected in a training domain ($\sourcedomain$) under a different set of uncertainties ($\mathbb{W}_\sourcedomain \subseteq \mathbb{W}_\targetdomain$, $\mathbb{V}_\sourcedomain \subseteq \mathbb{V}_\targetdomain$) resulting in a different distribution of training data. %

\noindent
\textbf{Tube and ancillary controller for \ac{DA}.}
To overcome these limitations, we design a \ac{DA} strategy that compensates for the effects of \textit{process and sensing uncertainties} encountered in $\targetdomain$. %
We do so by extending our previous approach \cite{tagliabue2021demonstration}, named \ac{SA}, which provided a strategy to efficiently learn a control policy (i.e., $\hat{\pi} : \mathfrak{R}^{n_x} \rightarrow \mathfrak{R}^{n_u}$) robust to \textit{process} uncertainty ($\mathbb{W}_\targetdomain$). \ac{SA} recognized that the tube in a \ac{RTMPC} \cite{mayne2005robust} represents a model of the states that the system may visit when subject to process uncertainties. \ac{SA} used the tube in \cite{mayne2005robust} to guide the selection of extra states for \ac{DA}, while the ancillary controller in \cite{mayne2005robust} provided a computationally efficient way to compute corresponding actions, maintaining the system inside the tube for every possible realization of the process uncertainty. 

\noindent
\textbf{Tube-NeRF.} Our new approach, named Tube-NeRF, employs the output feedback variant of \ac{RTMPC} presented in \cref{sec:tube_mpc}. This has two benefits: (i) the controller appropriately introduces extra conservativeness during demonstration collection to account for sensing uncertainties (via tightened constraint $\bar{\mathbb{X}}$ and $\bar{\mathbb{U}}$ in \cref{eq:ofrtmpc}); and (ii) the tube $\mathbb{Z}$ in \cref{eq:tube} additionally captures the effects of \textit{sensing} uncertainty, guiding the generation of extra observations for \ac{DA}. The new data collection and \ac{DA} procedure is as follows.

\subsubsection{Demonstration collection}
We collect demonstrations in $\sourcedomain$ using the output feedback \ac{RTMPC} \textit{expert} (\cref{sec:tube_mpc}).  
Each $T+1$-step demonstration $\bar{\tau}$ consists of: 
\begin{equation} \small
\bar{\tau} = \{(\mathbf{o}_t, \mathbf{u}_t, \usafe_t^*, \xsafe_t^*, \mathbf{X}^\text{des}_t, \xest_t) \mid t = 0, \dots, T \}.
\end{equation}

\subsubsection{Extra states and actions for synthetic data generation}
For every timestep $t$ in $\bar{\tau}$, we generate $N_{\text{synthetic},t} > 0$ (details on how $N_{\text{synthetic},t}$ is computed are provided in \cref{sec:tube_guided_real_images}) extra (state, action) pairs $(\mathbf{x}_{t,j}^+, \mathbf{u}_{t,j}^+)$, with $j=1,\dots,N_{\text{synthetic},t}$ by sampling extra states from the tube $\mathbf x_{t,j}^+ \in \xsafe_t^* \oplus \mathbb{Z}$, and computing the corresponding control action $\mathbf u_{t,j}^+$ using \cref{eq:ancillary_controller}:
\begin{equation} \small
\mathbf u_{t,j}^+ = \mathbf \usafe_t^* +  \mathbf K (\mathbf{x}_{t,j}^+ - \xsafe_t^*).
\label{eq:tubempc_feedback_policy}
\end{equation}
The resulting $\mathbf u_{t,j}^+$ is saturated to ensure that $\mathbf u_{t,j}^+ \in \mathbb{U}$.

\noindent
\subsubsection{Synthetic observations generation} \label{sec:synthetic_obs}
To generate the necessary data $\mathbf{o}_{t,j}^+ = (\mathcal{I}_{t,j}^+, \mathbf{o}_{\text{other},{t,j}}^+)$ input for the sensorimotor policy \cref{eq:policy} from the selected states $\mathbf{x}_{j,t}^+$, we employ observation models \cref{eq:obs_model} available for the expert. %
In the context of learning a visuomotor policy, we generate synthetic camera images $\mathcal{I}_{t,j}^+$ using an inverse pose estimator $\hat{g}^{-1}_{\text{\text{cam}}}$, mapping camera poses $\mathbf T_{IC}$ to images $\mathcal{I}$ via
$
\small
    \mathcal{I}_{t,j}^+ = \hat{g}_{\text{\text{cam}}}^{-1}(\mathbf T_{{IC}_{t,j}}^+),
$
where $\mathbf T_{IC}$ denotes a homogeneous transformation matrix from a world (inertial) frame $I$ to a camera frame $C$. $\hat{g}^{-1}_{\text{\text{cam}}}$ is obtained by generating a \ac{NeRF} of the environment (discussed in more details in \cref{sec:nerf_generation}) from the images $\mathcal{I}_0, \dots, \mathcal{I}_T$ in the collected demonstration $\bar{\tau}$, and by estimating the intrinsic/extrinsic of the camera onboard the robot. 
The camera poses $\textbf{T}_{{IC}_{t,j}}^+$ are obtained from the sampled states $\mathbf{x}_{t,j}^+$, which includes the robot's position and orientation. These are computed as $\textbf{T}_{{IC}_{t,j}}^+ = \textbf{T}_{IB}(\mathbf{x}_{t,j}^+) \hat{\textbf{T}}_{{BC}_{t,j}}$, where $\textbf{T}_{IB}$ is the transformation from the robot's body frame $B$ to the reference frame $I$, and $\hat{\textbf{T}}_{{BC}_{t,j}}$ are \b{perturbed transformation of the nominal camera extrinsic, where perturbations are introduced to accommodate uncertainties and errors in the extrinsics}.
Last, the full observations $\mathbf{o}_{t,j}^+$ are obtained by computing $\mathbf{o}_{\text{other},t,j}^+$, using \cref{eq:obs_model} and a selection matrix $\mathbf{S}$: 
\begin{equation} \small
    \mathbf {o}_{t,j}^+ = (\mathcal{I}_{t,j}^+, \mathbf{o}_{\text{other},t,j}^+), \; \; \mathbf{o}_{\text{other},t,j}^+ = \mathbf{S} \mathbf{C} \mathbf{x}_{t,j}^+.
\end{equation}

\subsubsection{Tube-guided selection of extra real observations}
\label{sec:tube_guided_real_images}
Beyond guiding the generation of extra synthetic data, we employ the tube of the expert to guide the selection of real-world observations from demonstrations ($\bar{\tau}$) for \ac{DA}. 
This procedure is useful at accounting for small imperfections in the NeRF and in the camera-to-robot extrinsic/intrinsic calibrations, further reducing the sim-to-real gap, and providing an avenue to ``ground'' the synthetic images with real-world data. 
This involves creating a database of the observations $\mathbf{o}$ in $\bar{\tau}$, indexed by the robot's estimated state $\xest$, and then selecting $N_{\text{real},t}$ observations at each timestep $t$ inside the tube ($\xest \in \xsafe^*_t \oplus \mathbb{Z}$), adhering to the ratio $N_{\text{real},t}/N_\text{samples} \leq \bar{\epsilon}$, where $0 < \bar{\epsilon} \leq 1$ is a user-defined parameter that balances the maximum ratio of real images to synthetic ones, and $N_\text{samples}$ is the desired number of samples (real and synthetic) per timestep. The corresponding action is obtained from the state associated with the image via the ancillary controller \cref{eq:tubempc_feedback_policy}. The required quantity of synthetic samples to generate, as discussed in Section \ref{sec:synthetic_obs}, is calculated using the formula $N_{\text{synthetic},t} = N_\text{samples} - N_{\text{real},t}$.

\subsubsection{Robustification to visual changes}
\label{sec:robustification_to_visual_changes}
To accommodate changes in brightness and environment, we apply several transformations to both real and synthetic images. These include solarization, adjustments in sharpness, brightness, and gamma, along with the application of Gaussian noise, Gaussian blur, and erasing patches of pixels using a rectangular mask.

\section{APPLICATION TO VISION-BASED FLIGHT}
In this section, we provide details on the design of the expert, the student policy, and the NeRF for agile flight. 
\label{sec:application_to_vision_flights}
\begin{figure}[t]
    \centering %
    \includegraphics[trim={140 10 60 5},clip, width=1\columnwidth]{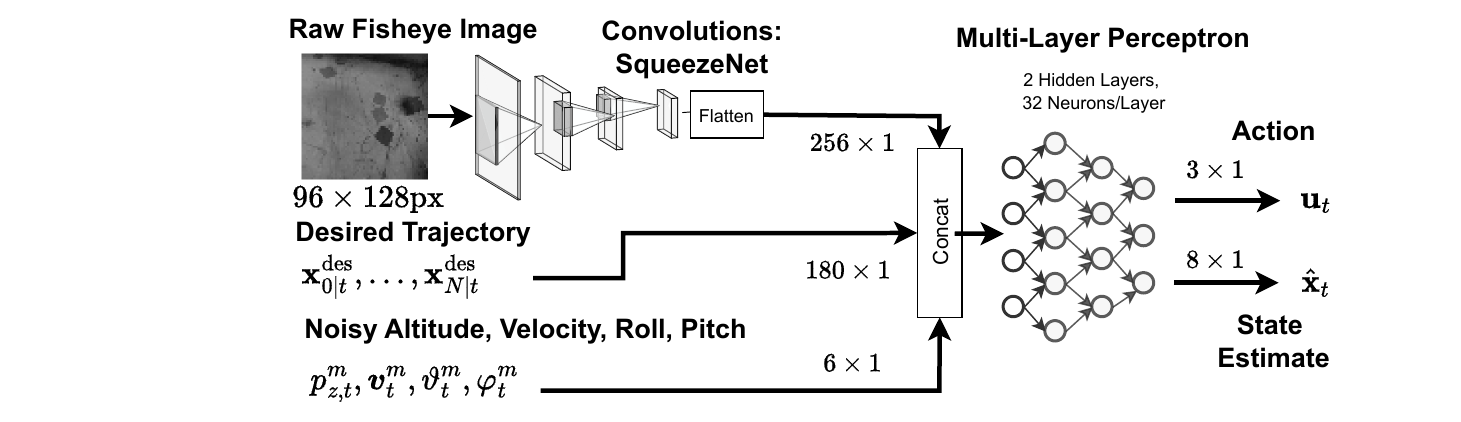}
    \caption{Architecture of the employed visuomotor student policy. The policy takes as input a raw camera image, a reference trajectory $\xdes_{0|t}, \dots, \xdes_{N|t}$ and noisy measurements of the altitude $p^m_z$, velocity $\boldsymbol{v}_t$ and tilt (roll $\varphi_t$, pitch $\vartheta_t$) of the multirotor. It outputs an action $\textbf{u}_t$, representing a desired roll, pitch, and thrust set-points for the cascaded attitude controller. The policy additionally outputs an estimate of the state $\xest_t$, which was found useful to promote learning of features relevant to position estimation.}
    \label{fig:policy_architecture}
    \vskip-4ex
\end{figure}

\textbf{Task.} 
We apply our framework to learn to track a figure-eight trajectory (lemniscate, with velocity up to $3.15$ m/s lasting $30$s), denoted \textbf{T1}. %
\textbf{Robot model.} The expert uses the hover-linearized model of a multirotor \cite{kamel2017linear}, with state
$
    \mathbf x = [ _I\boldsymbol p^T, _I \boldsymbol v^T,  _{\bar{I}} \varphi, _{\bar{I}} \vartheta ]^T,
$
(position $\boldsymbol{p} \in \mathfrak{R}^3$, velocity $\boldsymbol{v} \in \mathfrak{R}^3$, roll $\varphi \in \mathfrak{R}$, pitch $\vartheta \in \mathfrak{R}$, $n_x\!=\!8$), $I$ is an inertial reference frame, while ${\bar{I}}$ a yaw-fixed frame \cite{kamel2017linear}.
The control input $\mathbf{u}_t$ ($n_u\!=\!3$) is desired roll, pitch, and thrust, and it is executed by a cascaded attitude controller. 
\noindent
\textbf{Measurements.}
Our multirotor is equipped with a fisheye monocular camera, tilted $45 \deg$ downwards, that generates images $\mathcal{I}_t$ (size $128 \times 96$ pixels). In addition, we assume available onboard noisy altitude $_I p^m_z \in \mathfrak{R}$, velocity $_I \boldsymbol{v}^m \in \mathfrak{R}^3$ and roll $_I \varphi^m$, pitch $_I \vartheta^m$ and yaw measurements. This is a common setup in aerial robotics, where noisy altitude and velocity can be obtained, for example, via optical flow and a downward-facing lidar, while roll, pitch, and yaw can be computed from an IMU with a magnetometer, using a complementary filter \cite{euston2008complementary}. 
\noindent
\textbf{Student policy.} The student policy \cref{eq:policy}, whose architecture is shown in \cref{fig:policy_architecture}, takes as input an image $\mathcal{I}_t$ from the onboard camera, the reference trajectory $\mathbf X^\text{des}_t$, and 
$
\mathbf{o}_\text{other} \coloneqq [
_I p^m_z, 
{_I \boldsymbol{v}^m}^T, 
_{\bar{I}} \varphi^m, 
_{\bar{I}} \vartheta^m]^T,
$
and it outputs $\mathbf{u}_t$.  
A Squeezenet \cite{iandola2016squeezenet} is used to map $\mathcal{I}_t$ into a lower-dimensional feature space; it was selected for its performance at a low computational cost. %
To promote learning of internal features relevant to estimating the robot's state, the output of the policy is augmented to predict the current state $\xest$ (or $\mathbf{x}^+$ for the augmented data), modifying the training loss accordingly. This output was not used at deployment time, but it was found to improve the performance. 
\noindent \textbf{Output Feedback RTMPC and observer.} The expert uses the defined robot model for predictions, discretized with $T_s = 0.1$ s, and horizon $N = 30$, ($3.0$ s). $\mathbb{X}$ encodes safety and position limits, while $\mathbb{U}$ captures the maximum actuation capabilities. Process uncertainty in $\mathbb{W}_\targetdomain$ is assumed to be a bounded external force disturbance with magnitude \b{between to $10-19\%$} of the weight of the robot and random direction, close to the physical limits of the platform, \b{and the tube of the expert is computed assuming $\bar{\mathbb{W}}_\targetdomain$ equal to $20\%$ of the weight of the robot}.
The state estimator \cref{eq:estimator} is designed by using as measurement model in \cref{eq:obs_model}:
$
    \bar{\mathbf{o}}_t = \mathbf x_t + \mathbf v_t
$
where we assume 
$\mathbf v_t\!\sim\!\mathcal {N}(\mathbf{0}_{n_x}, [\boldsymbol{\sigma}^T_\text{cam},\boldsymbol{\sigma}^T_\text{other}]^T)$. 
We therefore set $\mathbb{V}_\targetdomain = \{ \mathbf{v}_t \; | \; \|\mathbf v\|_\infty \leq 3 \left[ \boldsymbol{\sigma}_{\text{cam}}^T, \boldsymbol{\sigma}_\text{other}^T \right]^T \}$, with
$3\boldsymbol{\sigma}_{\text{cam}} = [0.6, 0.6]^T$ (units in $m$) and 
$3\boldsymbol{\sigma}_\text{other} = [0.4, 0.2, 0.2, 0.2, 0.05, 0.05]^T$ (units in $m$ for altitude, $m/s$ for velocity, and $rad$ for tilt). These conservative but realistic parameters are based on prior knowledge of the worst-case performance of vision-based estimators in our relatively feature-poor flight space.
The observer gain matrix $\mathbf L$ is computed by assuming fast state estimation dynamics, (poles of $\mathbf{A - LC}$ at $30.0$ rad/s).
\vskip-10pt

\subsubsection{Procedure to generate the NeRF of the environment}
\label{sec:nerf_generation}
\begin{inparaenum}[(i)]
\item \textbf{Dataset:} A \ac{NeRF} of the environment, the MIT-Highbay, is generated from  about $100$ images collected during a single real-world demonstration of the figure-eight trajectory (\textbf{T1}) intended for learning, utilizing full-resolution images ($640\times480$ pixels) from the fisheye camera onboard the Qualcomm Snapdragon Flight Pro board of our UAV.
\item \textbf{Extrinsic/Intrinsic:} The extrinsic and intrinsic parameters of the camera are estimated from the dataset using structure-from-motion (COLMAP \cite{schoenberger2016sfm}, RADTAN camera model). %
\item \textbf{Frame Alignment:} The scale and homogeneous transformation aligning the reference frame used by the COLMAP with the reference frame used by the robot's state estimator are determined via the trajectory alignment tool EVO \cite{grupp2017evo}. This enables the integration of the NeRF as an image rendering tool in a simulation/\ac{DA} framework.
\item \textbf{NeRF Training:} Instant-NGP \cite{muller2022instant} is utilized to train the \ac{NeRF}. The scaling of the Axis-Aligned Bounding Box used by the Instant-NGP is manually adjusted to ensure that the reconstruction is photorealistic in the largest possible volume.
\item \textbf{Images rendering for \ac{DA}:} Novel images are rendered using the same camera intrinsics identified by COLMAP. The camera extrinsics, mapping from the robot's IMU to the optical surface, are determined via Kalibr \cite{furgale2013unified}, using an ad-hoc dataset. 
An example of an image from the NeRF is shown in \cref{fig:approach_overview}.
\end{inparaenum}

\section{Evaluation}
\subsection{Evaluation in simulation}

Here we numerically evaluate the efficiency (training time, number of demonstrations), robustness (average episode length before violating a state constraint, success rate) and performance (\textit{expert gap}, the relative error between the stage cost $\sum_{t} \mathbf \|\mathbf{e}_t\|_{\mathbf Q}^2+ \| \mathbf {u}_t \|^2_{\mathbf R}$ of the expert and the one of the policy) of Tube-NeRF. \b{Note that the number of demonstrations provides a metric useful not only to estimate real-world data collection efforts, but also the number of environment interactions required in simulation which, depending on the simulation environment considered, may be computationally costly (e.g., in fluid-dynamic simulations)}. We use PyBullet to simulate realistic full nonlinear multirotor dynamics \cite{kamel2017linear}, rendering images using the NeRF obtained in \cref{sec:nerf_generation} -- combined with realistic dynamics, the NeRF provides a convenient framework for training and numerical evaluations of policies. The considered task consists of following the figure-eight trajectory \textbf{T1} (lemniscate, length: $300$ steps) used in \cref{sec:nerf_generation}, starting from $\mathbf{x}_0 \! \sim \!\text{Uniform}(-0.1, 0.1) \in \mathbb{R}^8$, without violating state constraints. The policies are deployed in two target environments, one with sensing noise affecting the measurements $\mathbf{o}_\text{other}$ (the noise is Gaussian distributed with parameters as defined in \cref{sec:application_to_vision_flights}) and one that additionally presents wind disturbances, sampled from $\mathbb{W}_\targetdomain$ (also with bounds as defined in \cref{sec:application_to_vision_flights}). 
\textbf{Method and baselines.}
We apply Tube-NeRF to BC and DAgger, comparing their performance without any \ac{DA}; Tube-NeRF$-N_\text{samples}$, with $N_\text{samples} = \{50, 100\}$, denotes the number of observation-action samples generated for every timestep by uniformly sampling states inside the tube. We additionally combine BC and DAgger with \ac{DR} by applying, during demonstration collection, an external force disturbance sampled from $\mathbb{W}_\targetdomain$. We set $\beta$, the hyperparameter of DAgger controlling the probability of using actions from the expert instead of the learner policy, to be $\beta=1$ for the first set of demonstrations, and $\beta=0$ otherwise. 

\begin{figure}[t]
\centering
\begin{subfigure}[b]{0.45\textwidth}
  \includegraphics[trim={5 5 5 7}, clip, width=\columnwidth]{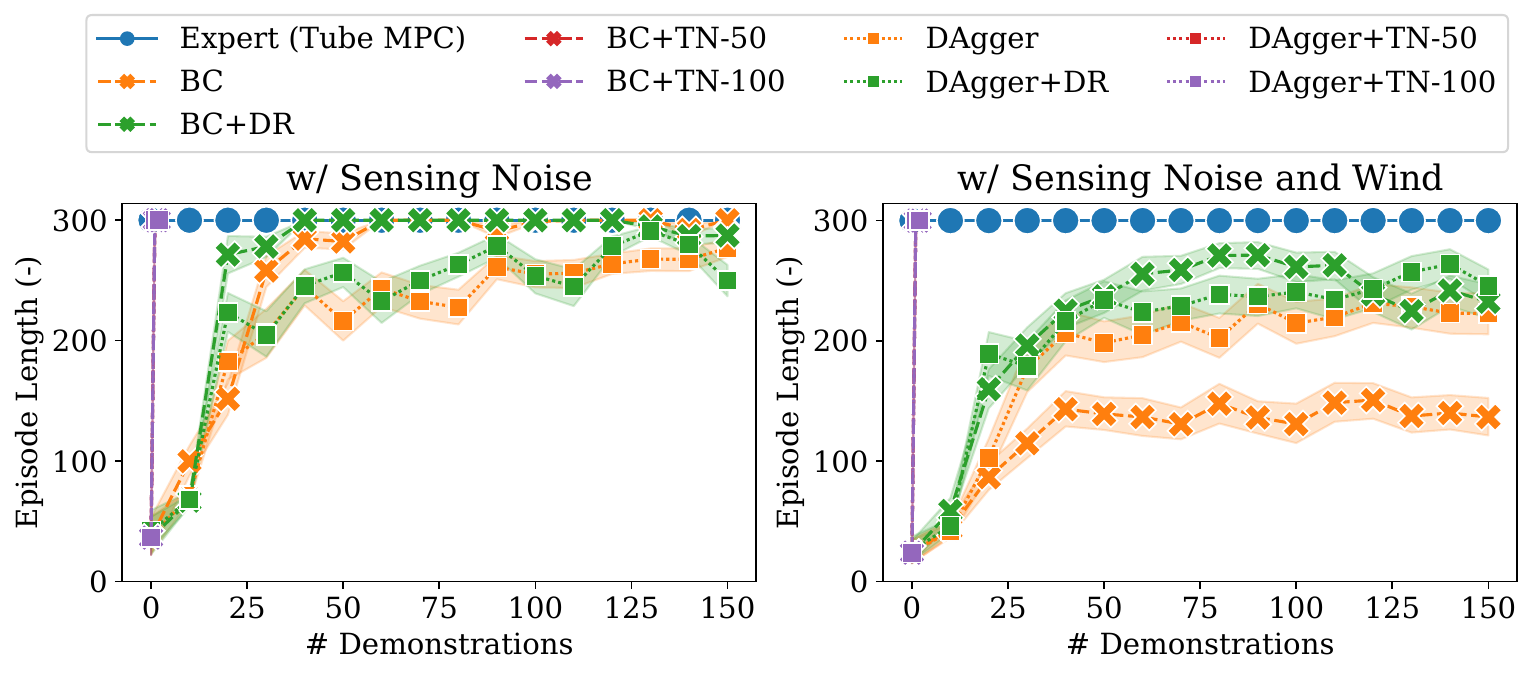}
\end{subfigure}
\begin{subfigure}[b]{0.45\textwidth}
  \includegraphics[trim={5 5 5 75}, clip, width=\columnwidth]{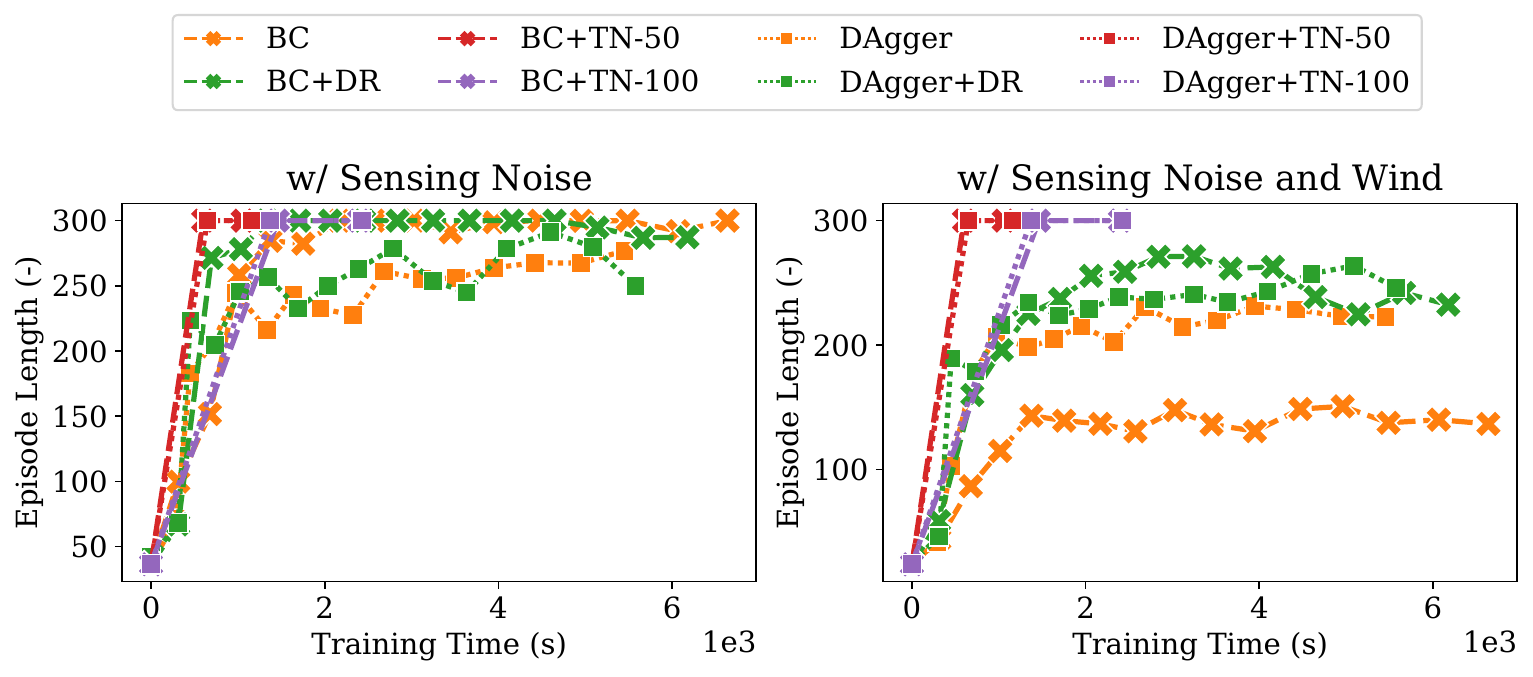}
\end{subfigure}
\caption{Episode length \b{(timestep before a state constraint violation, up to $300$) vs. number of demonstrations collected from the expert, and vs }the training time (the time required to collect such demonstrations in simulation, and to train the policy). 
\b{This shows} that policies trained with Tube-NeRF (TN) archive full episode length after a single demonstration, and require less than half of the training time than the best-performing baselines (DR-based methods). 
Note that the lines of Tube-NeRF-based approaches vs the number of demonstrations overlap. Shaded areas are $95\%$ confidence intervals. Note that to focus our study on the effects of process uncertainties and sensing noise, we do not apply visual changes to the environment, nor the robustification to visual changes (\cref{sec:robustification_to_visual_changes}). Evaluations across $10$ seeds, $10$ times per seed.}
\label{fig:tube_nerf_numerical_analsys}
\vskip-2ex
\end{figure}

\begin{table}[t]
    \caption{robustness, performance, and demonstration efficiency of \ac{IL} methods for visuomotor policy learning. 
    An approach \textit{easy} if it does not require disturbances during the demonstration collection, and it is \textit{safe} if it does not violate state constraints (e.g., wall collision) during training. 
    \textit{Success Rate} is the percentage of trajectories that do no violate state constraints. 
    \textit{Performance} is the relative error between the cost of the trajectory generated by output feedback RTMPC (expert) and the ones from the policy. 
    \textit{Demonstration efficiency} is the number of demonstrations required to achieve at least $70\%$ \textit{success rate}. Performance and robustness of the baselines are evaluated after $150$ demonstrations, while Tube-NeRF methods  after \textit{only} $2$ demonstrations.}
\newcolumntype{P}[1]{>{\centering\arraybackslash}p{#1}}
    \vskip-1ex
    \tiny
    \renewcommand{\tabcolsep}{1pt}
    \centering
    \begin{tabular}{|p{0.9cm}p{0.9cm}||P{0.7cm}|P{0.7cm}|P{0.7cm}|P{0.7cm}|P{0.7cm}|P{0.7cm}|P{0.7cm}|P{0.7cm}|}
    \hline
    \multicolumn{2}{|c||}{Method} &  
    \multicolumn{2}{c|}{Training} & 
    \multicolumn{2}{c|}{\makecell{Robustness\\succ. rate (\%)}} &
    \multicolumn{2}{c|}{\makecell{Performance\\expert gap (\%)}}  &
    \multicolumn{2}{c|}{\makecell{Demonstration\\Efficiency}}\\
    
           Robustif.       & Imitation   & Easy          & Safe             & Noise                 &  Noise, Wind      &  Noise             &  Noise, Wind      & Noise             &  Noise, Wind \\
        \hline
        \hline
        \multirow{2}{*}{-} & BC          & \textbf{Yes}  & \textbf{Yes}     & 95.5                    & 13.0              & 83.1               & 21.3              &     30           &  - \\
                           & DAgger      & \textbf{Yes}  & No               & 60.5                    & 45.5              & 138.6              & 43.0              &     -            &  - \\ 
        \hline
        \multirow{2}{*}{DR}& BC          & No            & \textbf{Yes}     & 80.0                    & 46.5              & 72.5               & 59.2              &     20           &  90 \\ 
                           & DAgger      & No            & No               & 67.0                    & 62.0              & 82.2               & 59.6              &     90           &  - \\
        \hline
        \multirow{2}{*}{TN-100} & BC     & \textbf{Yes}  & \textbf{Yes}     & \textbf{100.0}          & \textbf{100.0}             & \textbf{8.3}       & \textbf{9.1}      & \textbf{1}        & \textbf{1} \\
                                & DAgger & \textbf{Yes}  & \textbf{Yes}     & \textbf{100.0}          & \textbf{100.0}             & 14.4               & 9.7               & \textbf{1}        & \textbf{1} \\
        \hline
        \multirow{2}{*}{TN-50} & BC      & \textbf{Yes}  & \textbf{Yes}     & \textbf{100.0}          & \textbf{100.0}             &  18.4              & 11.8              & \textbf{1}        & \textbf{1} \\
                               & DAgger  & \textbf{Yes}  & \textbf{Yes}     & \textbf{100.0}          & \textbf{100.0}             &  15.4              & 11.2              & \textbf{1}        & \textbf{1} \\
        \hline
    \hline
    \end{tabular}%
     \label{tab:analysis_at_convergence}
     \vskip-4ex
\end{table}

\noindent
\textbf{Evaluation details.}
For every method, we:
\begin{inparaenum}[(i)]
    \item collect $K$ new demonstrations ($K=1$ for Tube-NeRF, $K=10$ otherwise) via the output feedback RTMPC expert and the state estimator;
    \item update\footnote{Policies are trained for $50$ epochs with the ADAM optimizer, learning rate $0.001$, batch size $32$, and terminating training if the loss does not decrease within $7$ epochs. Tube-NeRF uses only the newly collected demonstrations and the corresponding augmented data to update the previously trained policy.} a student policy using all the demonstrations collected so far; 
    \item evaluate the obtained policy in the considered target environments, for $10$ times each, starting from different initial states;
    \item repeat from (i).
\end{inparaenum}
\b{Note that in our comparison the environment steps at its highest possible rate (simulation time is faster than wall-clock time), providing an advantage in terms of data collection time to those methods that require collecting a large number of demonstrations (our baselines).}
\noindent \textbf{Results.} %
\cref{fig:tube_nerf_numerical_analsys} highlights that all Tube-NeRF methods, combined with either DAgger or BC, can achieve complete robustness (full episode length) under combined sensing and process uncertainties after a single demonstration. The baseline approaches require $20$-$30$ demonstrations to achieve a full episode length in the environment without wind, and the best-performing baselines (methods with \ac{DR}) require about $80$ demonstrations to achieve their top episode length in the more challenging environment. Similarly, Tube-NeRF requires less than \textit{half} training time than \ac{DR}-based methods to achieve higher robustness in this more challenging environment, and reducing the number of samples (e.g, Tube-NeRF-$50$) can further improve the training time. \b{The time to generate the NeRF, not shown in Fig. 4, was approximately $5$ minutes ($20000$ epochs on an \texttt{RTX 3090 GPU}). Even accounting for this time, TN is significantly faster than collecting real-world demonstrations (the $80$ demonstrations required by DR correspond to $40$ minutes of real-world data, followed by the time to train the policy). In addition, if the NeRF is combined with a simulation of the dynamics of the robot (creating a photo-realistic simulator), our DA strategy still provides benefits in terms of performance and training time.}. \cref{tab:analysis_at_convergence} additionally highlights the small gap of Tube-NeRF policies from the expert in terms of tracking performance (\textit{expert gap}), and shows that increasing the number of samples (e.g., Tube-NeRF-$100$) benefits performance.

\subsection{Flight experiments}
\begin{table*}
    \vskip 2ex
    \renewcommand{\arraystretch}{1.4}
    \caption{Position \ac{RMS} Tracking Errors
    when following a figure-eight (lemniscate) trajectory (T1) and a circular trajectory (T2). This  highlights the low errors of the sensorimotor policy, comparable to the ones of the expert whose position is estimated via a motion capture system. We apply wind disturbances with a leaf blower (With wind), and extra sensing noise in the altitude, velocity and orientation input of the policy (High noise). %
    Note that T2 as been obtained without collecting corresponding images from a real-world demonstration, and heavily relies on NeRF data (see \cref{fig:tube_guided_sampling}).
    Experiments repeated $3$ times, except T2 with wind low noise ($2$ times), and slung-load (once).}
    \vskip-1ex
    \newcolumntype{P}[1]{>{\centering\arraybackslash}p{#1}}
    \centering
    \scriptsize
    \resizebox{0.75\textwidth}{!}{
    \renewcommand{\tabcolsep}{5.5pt}
    \begin{tabular}{c||c|c|c|c|c|c|c||c|c|c|c||}
    \hline
    \multicolumn{1}{c||}{} & \multicolumn{7}{c||}{\textbf{RMSE (m, \( \downarrow \)) for T1: Lemniscate (30s, real-world demo.)}} & \multicolumn{4}{c||}{\textbf{RMSE (m, \( \downarrow \)) for T2: Circle (30s, No real-world demo.)}} \\
    \multicolumn{1}{c||}{} & \multicolumn{2}{c|}{\textbf{Expert + Motion Capture}} & \multicolumn{5}{c||}{\textbf{Student}} & \multicolumn{4}{c||}{\textbf{Student}} \\
    \multicolumn{1}{c||}{} & \multicolumn{1}{c|}{No wind} & \multicolumn{1}{c|}{With wind} & \multicolumn{2}{c|}{No wind} & \multicolumn{2}{c|}{With wind} & \multicolumn{1}{c||}{\parbox{0.7cm}{\centering Slung\\load}} & \multicolumn{2}{c|}{No wind} & \multicolumn{2}{c||}{With wind} \\
     & Low noise & Low noise & Low noise & High noise  & Low noise & High noise &  & Low noise & High noise &  Low noise & High noise \\
    \hline
    \( x \) & 0.19 \( \pm \) 0.00 & 0.20 \( \pm \) 0.01 & 0.30 \( \pm \) 0.01 & 0.33 \( \pm \) 0.00 & 0.28 \( \pm \) 0.02 & 0.40 \( \pm \) 0.01 & 0.21 & 0.33 \( \pm \) 0.01 &  0.34 \( \pm \) 0.01 & 0.33 \( \pm \) 0.00 & 0.34 \( \pm \) 0.04 \\
    \( y \) & 0.17 \( \pm \) 0.01 & 0.21 \( \pm \) 0.01 & 0.16 \( \pm \) 0.01 & 0.21 \( \pm \) 0.01 & 0.26 \( \pm \) 0.01 & 0.22 \( \pm \) 0.03 & 0.44 & 0.16 \( \pm \) 0.01 &  0.27 \( \pm \) 0.01 & 0.22 \( \pm \) 0.03 & 0.28 \( \pm \) 0.01 \\
    \( z \) & 0.12 \( \pm \) 0.01 & 0.13 \( \pm \) 0.02 & 0.11 \( \pm \) 0.01 & 0.20 \( \pm \) 0.02 & 0.17 \( \pm \) 0.01 & 0.17 \( \pm \) 0.01 & 0.06 & 0.11 \( \pm \) 0.08 &  0.15 \( \pm \) 0.06 & 0.07 \( \pm \) 0.00 & 0.14 \( \pm \) 0.02 \\
    \hline
    \end{tabular}
    } %
    \label{tab:rmse_all_tracking}
    \vskip-3ex
\end{table*}

\begin{figure*}[ht!]
\captionsetup[sub]{font=footnotesize}
\centering
\begin{subfigure}{0.245\textwidth}
    \centering
    \includegraphics[trim={40, 40, 10, 40}, clip,width=1.0\textwidth]{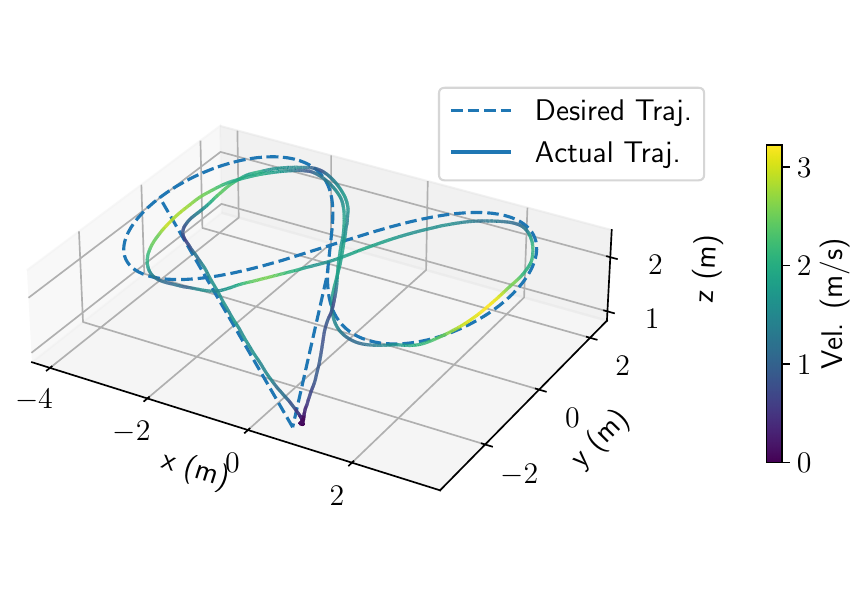}
    \caption{T1, student w/o wind.}
    \vskip-2ex
    \label{fig:t1_no_wind}
\end{subfigure}%
\vspace{0.1cm}
\begin{subfigure}{0.245\textwidth}
    \centering
    \includegraphics[trim={10, 40, 10, 40}, clip, width=1.0\textwidth]{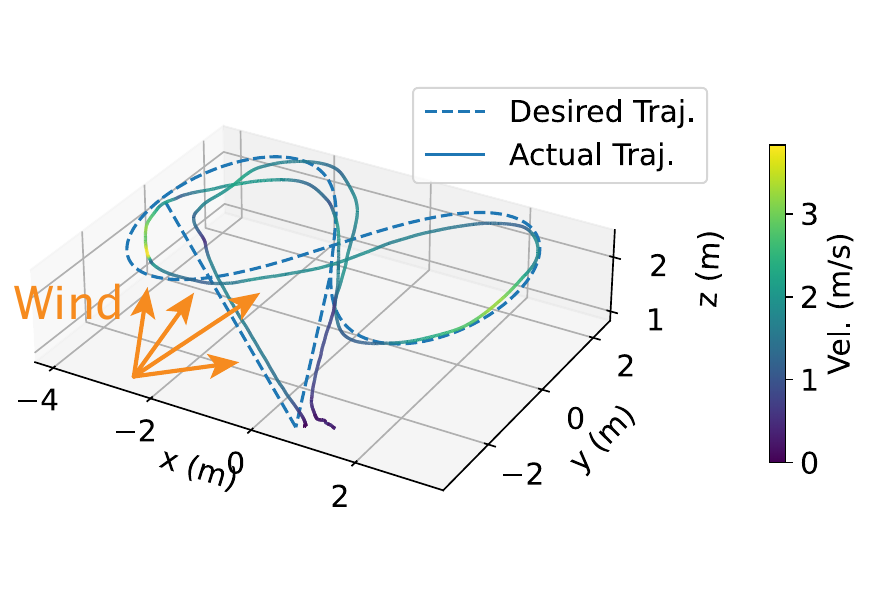}
    \caption{T1, student w/ wind.}
    \vskip-2ex
    \label{fig:t1_with_wind}
\end{subfigure}%
\vspace{0.1cm}
\begin{subfigure}{0.245\textwidth}
    \centering
    \includegraphics[trim={10, 20, 10, 30}, clip, width=1.0\textwidth]{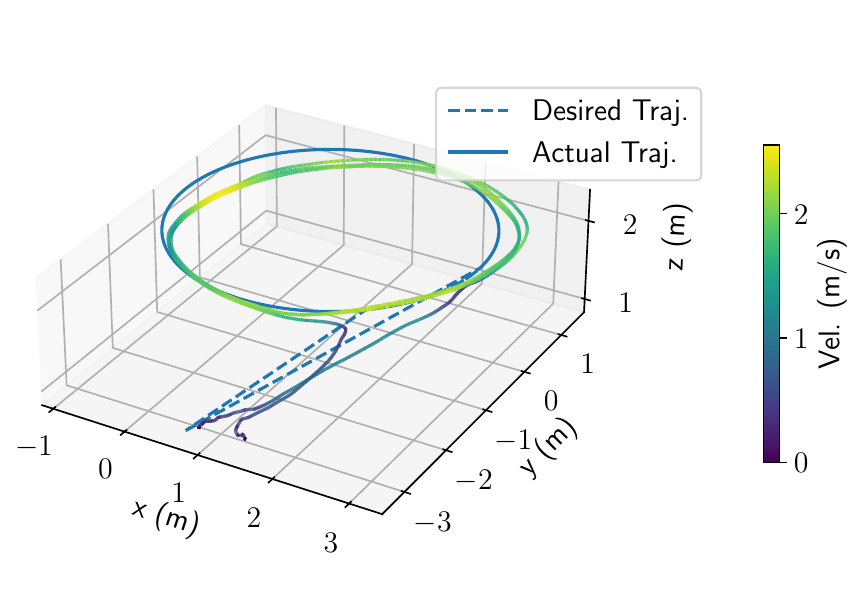}
    \caption{T2, student  w/o wind.}
    \vskip-2ex
    \label{fig:t2_no_wind}
\end{subfigure}%
\vspace{0.1cm}
\begin{subfigure}{0.245\textwidth}
    \centering
    \includegraphics[trim={10, 20, 10, 30}, clip, width=1.0\textwidth]{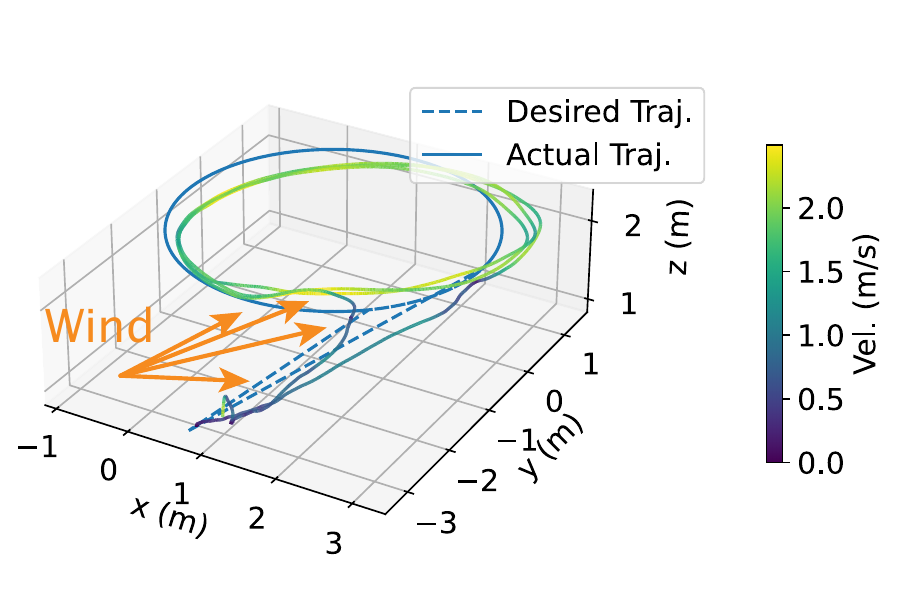}
    \caption{T2, student w/ wind.}
    \vskip-2ex
    \label{fig:t2_with_wind}
\end{subfigure}
    \vskip-0.5ex
    \caption{\b{\textbf{Qualitative evaluation in experiments}, highlighting the high velocity and the challenging 3D motion that the student policy can execute under uncertainties}. 
    }
    \label{fig:one_traj_performance}
    \vskip-2ex
\end{figure*}

\begin{figure}
    \centering
    \includegraphics[trim={0, 300, 0, 0}, clip, width=1\columnwidth]{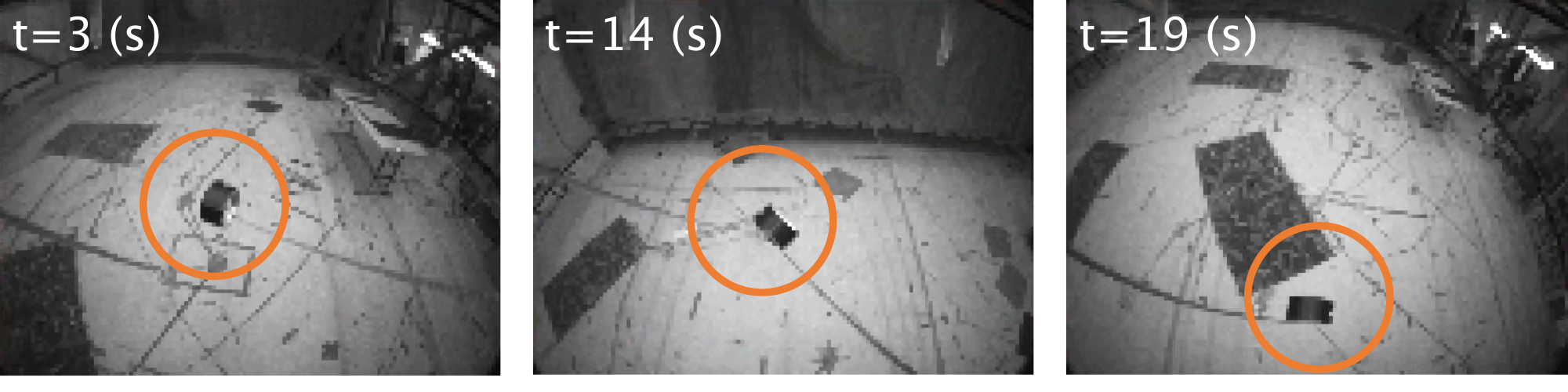}\vskip-0.5ex
    \caption{
    \textbf{Robustness to visual uncertainties}: we show images captured by the onboard camera while using the proposed sensorimotor policy for localization and tracking of \textbf{T1}, with a slung load (tape roll, $0.2$ Kg) attached to the robot. The slung load (circled) repeatedly enters the field of view of the onboard camera, without however compromising the success of the maneuver. \b{We hypothesize that randomly deleting patches of pixels during training (Sec. \ref{sec:robustification_to_visual_changes}) contributes to achieving robustness to this disturbance.}}
    \label{fig:slung_load_onboard_images}
    \vskip-2ex
\end{figure}
\begin{table}[t]
\small
    \renewcommand{\arraystretch}{1.5}
    \caption{\textbf{Time (ms) required to generated a new action for the output feedback RTMPC only (Expert) and the proposed sensorimotor \ac{NN} policy (Policy)}. Our  policy is $\mathbf{9.8}$ faster (offboard) and $\mathbf{5.6}$ times faster (onboard) than the expert. We note that the computational cost of the expert is only based on computing actions from states, while the policy additionally performs localization (actions from images); therefore the reported data represents a \textbf{lower bound} on the cost reductions introduced by the policy. For a fair comparison, the onboard expert uses an highly-optimized C/C++ implementation.
    The offboard computer, used for our numerical evaluation and training, is an \texttt{AMD Threadripper 3960X} with two {RTX 3090}. The onboard implementation (optimized for speed for both the policy and the expert) uses an \texttt{NVIDIA Jetson TX2}. }
    \vskip-2ex
    \label{tab:computational_cost}    \centering
    \resizebox{1.0\columnwidth}{!}{
    \begin{tabular} {|C{1.2cm}| C{2.7cm} |C{2.0cm}|C{0.6cm}|C{0.6cm}|C{0.6cm}|C{0.6cm}| }
    \multicolumn{3}{c}{} & \multicolumn{4} {c}{Time (ms)} \\
    \hline
        \textbf{Computer} & \textbf{Method}  & \textbf{Setup} & \textbf{Mean} & \textbf{SD} & \textbf{Min} & \textbf{Max}\\
        \hline 
        \hline 
        \multirow{2}{*}{Offboard} & Expert (MPC only)          & CVXPY/OSQP &  $30.3$ & $41.5$ & $4.8$ & $244$  \\
        \cline{2-7}
        & \textbf{Policy} (MPC+vision)  & PyTorch                       & $\mathbf{3.1}$ & $\mathbf{0.0}$ & $\mathbf{0.8}$ & $\mathbf{1.0}$  \\
        \hline
        \hline
        \multirow{2}{*}{Onboard} & Expert (MPC only)           & CVXGEN &  $8.4$ & $1.4$ & $4.5$ & $15.9$  \\
        \cline{2-7}
        & \textbf{Policy} (MPC+vision) & ONNX/TensorRT  &  $\mathbf{1.5}$ & $\mathbf{0.2}$ & $\mathbf{1.4}$ & $\mathbf{9.2}$  \\
        \hline
    \end{tabular}
    }
\vskip-3ex
\end{table}

\begin{figure}[t!]
    \centering
    \includegraphics[trim={0.1cm 0.58cm 0 0.2cm},clip, width=1\columnwidth]{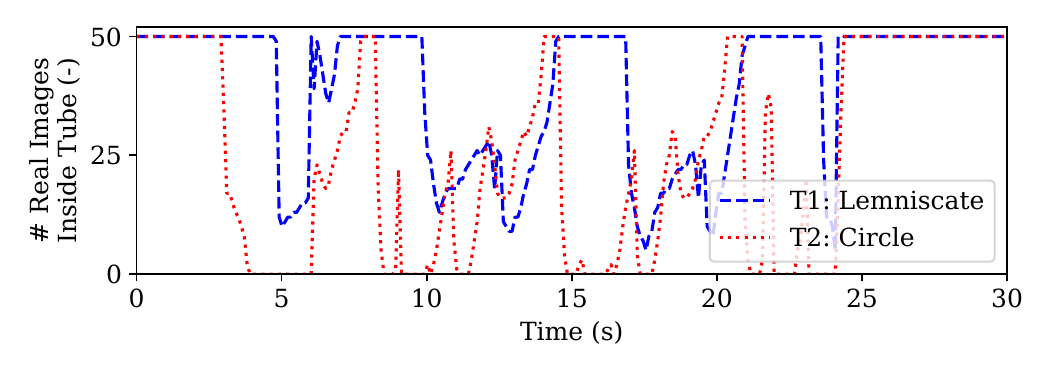} \vskip-0.5ex
    \caption{Number of real images sampled from the tube to perform data augmentation using \texttt{Tube-NeRF-100}, as a function of time in the considered trajectory. The considered trajectories are a Lemniscate trajectory (T1), which is the same as the one executed for real-world data collection, and a Circle trajectory (T2), which is different from the one executed for data collection. The results highlight that (1) the tube can be used to guide the selection of real-world images for data augmentation and (2) the synthetic images from the NeRF are a key component of our data augmentation strategy, as there are multiple segments of the circular trajectory (T2) where no real images are presents, but the sensorimotor policy successfully controls the robot in the real-world experiments.}
    \label{fig:tube_guided_sampling}
    \vskip-4ex
\end{figure}
\begin{figure}
    \centering
    \begin{tikzpicture}
    \node[inner sep=0pt] (figure) at (0,0) {\includegraphics[width=\columnwidth]{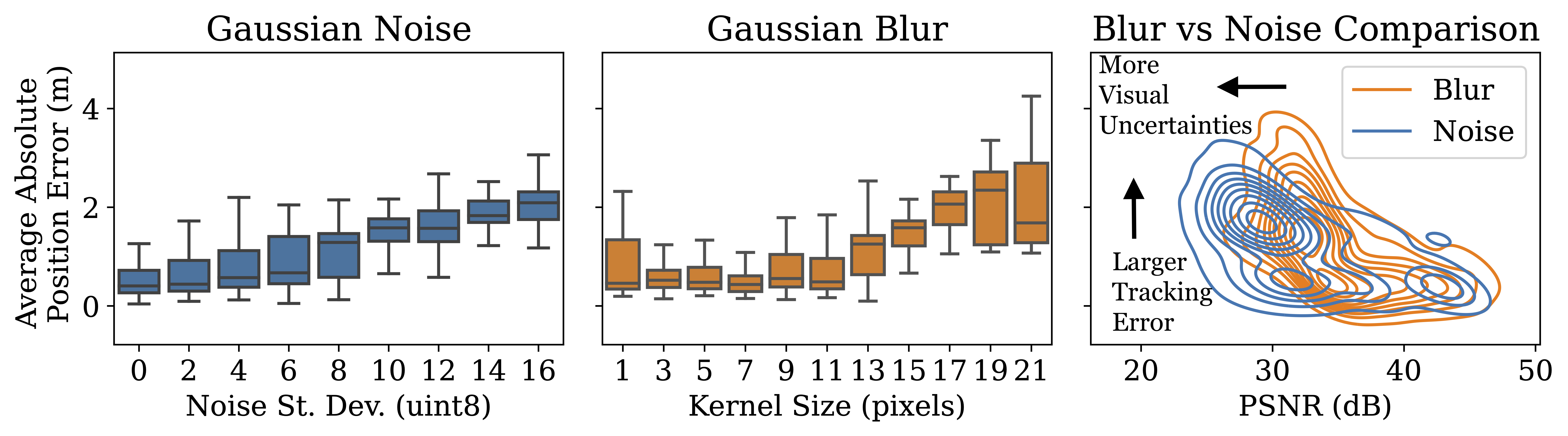}};
    \end{tikzpicture}
    \caption{\b{Position tracking error under wind disturbances for T1 in simulation as a function of different types and magnitude of noise in the visual input. PSNR is the Peak Signal to Noise Ratio, and lower PSNR denotes a larger amount of noise corrupting the image. The maximum noise level shown corresponds to to $0\%$ success rate. Contour lines in the Comparison plot show the density of the errors. }}
    \label{fig:visual_input_ablation}
    \vskip-4ex
\end{figure}

We now validate the data efficiency of Tube-NeRF highlighted in our numerical analysis by evaluating the obtained policies in real-world experiments. We do so by deploying them on an NVIDIA Jetson TX2 (at up to $200$ Hz, TensorRT) on the MIT-ACL multirotor. The policies take as input the fisheye images generated at $30$ or $60$ Hz by the onboard camera. The altitude, velocity and roll/pitch inputs that constitute $\textbf{o}_{\text{other},t}$ are, for simplicity, obtained from the onboard estimator (a filter fusing IMU with poses from a motion capture system), corrupted with additive noise (zero-mean Gaussian, with parameters as defined in \cref{sec:application_to_vision_flights}) in the scenarios denoted as ``high noise''. We remark that no information on the horizontal position of the UAV is provided to the policy, and horizontal localization must be performed from images. 
We consider two tasks, tracking the lemniscate trajectory (\textbf{T1}), and tracking a new circular trajectory (denoted \textbf{T2}, velocity up to $2.0$ m/s, duration of $30$ s). No real-world images have been collected for \textbf{T2}, therefore this task is useful to stress-test the novel-view synthesis abilities of the approach, using the NeRF and the nonlinear simulated robot dynamics as a simulation framework. \textbf{Training.} We train one policy for each task, using a single task demonstration collected with \texttt{DAgger+Tube-NeRF-100} in our NeRF-based simulated environment. During \ac{DA}, we try to achieve an equal amount between synthetic images (from the \ac{NeRF}) and real ones (from the database), setting $\bar{\epsilon} = 0.5$. \cref{fig:tube_guided_sampling} reports the number of sampled real images from the database, highlighting that the tube is useful at guiding the selection of real images, but that synthetic images are a key part of the DA strategy, \b{as \textbf{T2} presents multiple parts without any real image available}.
\textbf{Performance under uncertainties.} \cref{fig:one_traj_performance} and \cref{tab:rmse_all_tracking} show the trajectory tracking performance of the learned policy under a variety of real-world uncertainties. Those uncertainties include (i) model errors, such as poorly known drag and thrust to voltage mappings; (ii) wind disturbances, applied via a leaf-blower, (iii) sensing uncertainties (additive Gaussian noise \b{to the partial state measurements}), and %
(iv) visual uncertainties, produced by attaching a slung-load that repeatedly enters the field of view of the camera, as shown in \cref{fig:slung_load_onboard_images}. 
These results highlight that (a) policies trained after a single demonstration collected in our NeRF-based simulator using Tube-NeRF are robust to a variety of uncertainties while maintaining tracking errors comparable to the ones of the expert (\cref{tab:rmse_all_tracking}, \cref{fig:ofrtmpc_example}), while reaching velocities up to $3.5$ m/s, and even though the expert localizes using a motion capture system, while the policy uses images from the onboard camera to obtain its horizontal position. In addition, (b) our method enables learning of vision-based policies for which no real-world task demonstration has been collected, \b{effectively acting as a simulation framework}, as shown by the successful tracking of \textbf{T2}, which relied entirely on synthetic training data for large portions of the trajectory (\cref{fig:tube_guided_sampling}), and was obtained using a single demonstration in the NeRF-based simulator. Due to the limited robustness achieved in simulation (\cref{tab:analysis_at_convergence}), we do not deploy the baselines on the real robot. \textbf{Efficiency at deployment and latency.} \cref{tab:computational_cost} shows that onboard the policy requires on average only $\textbf{1.5}$ ms to compute a new action from an image, being at least $\mathbf{5.6}$ faster than a highly-optimized (C/C++) expert. Note that the reported computational cost of the expert is based on the cost of control only (no state estimation), therefore the actual computational cost reduction provided by the policy is even larger. \b{Online, image capture (independent of our method, nominal light conditions) has a latency of about $15$ms, while image pre-processing and transfer to the TX2 takes less than $2$ms.}. \b{
\textbf{Sensitivity to uncertainties in the visual input}. We study the closed-loop performance of the policy under different types of uncertainties in the visual input by monitoring the position tracking error (in simulation, tracking T1, under wind) as a function of different types and magnitudes of noise applied to the images rendered by the NeRF and input to the policy. Specifically, we consider 1) Gaussian noise, representing the presence of high-frequency disturbances/uncertainties such as new visual features in the environment, and 2) Gaussian Blur, capturing visual changes that reduce high-frequency features, such as the ones caused by weather changes, and/or the effects of using a low-quality NeRF for training. 
The results, shown in \cref{fig:visual_input_ablation}, highlight that 1) the visual input plays a key role in the output of the policy and, while our approach is not specifically designed for environments with rapidly changing visual appearance, it also shows 2) the overall robustness of the policy to visual uncertainties. Last, the comparison of the effects of the two types of noise highlights that 3) the policy has lower sensitivity to the presence of new high-frequency visual uncertainties, as the position errors grow slower when more Gaussian Noise is applied (lower PSNR) than for Gaussian Blur.
}

\section{DISCUSSION AND CONCLUSIONS}
We have presented Tube-NeRF, a strategy for efficient IL of robust end-to-end visuomotor policies \b{that achieve real-world robustness in trajectory tracking from images on a multirotors. Tube-NeRF leverages a robust controller, output feedback \ac{RTMPC}, to collect demonstrations that account for process and sensing uncertainties. In addition, properties of the controller are used to guide a data augmentation procedure, using a combination of a database of real-world images, a NeRF of the environment, and randomization procedures in image space to obtain novel relevant views. For each extra sensorial input, corresponding actions are \textit{efficiently} computed  using an ancillary controller, an integral part of the control framework.} We have tailored our approach to localization and control of an aerial robot, showing in numerical evaluations that Tube-NeRF can learn a robust visuomotor policy from a single demonstration, outperforming \ac{IL} baselines in demonstration and computational efficiency. Our experiments have validated the numerical finding, achieving accurate trajectory tracking using an onboard policy ($1.5$ ms average inference time) that relied entirely on images to infer the horizontal position of the robot, despite challenging 3D motion and uncertainties. In future work, \b{we aim to generalize the approach to dynamic environments by further randomization in image/NeRF space (using \cite{Li2023ClimateNeRF} for weather conditions), and use event-cameras to ensure performance in poorly-lit environments}. %

\balance
\begin{spacing}{0.9}
\bibliographystyle{IEEEtran}

\bibliography{root_shortened}
\end{spacing}
\balance
\begin{acronym}
\acro{OC}{Optimal Control}
\acro{LQR}{Linear Quadratic Regulator}
\acro{MAV}{Micro Aerial Vehicle}
\acro{GPS}{Guided Policy Search}
\acro{UAV}{Unmanned Aerial Vehicle}
\acro{MPC}{model predictive controller}
\acro{RTMPC}{robust tube model predictive controller}
\acro{DNN}{deep neural network}
\acro{BC}{Behavior Cloning}
\acro{DR}{Domain Randomization}
\acro{SA}{Sampling Augmentation}
\acro{IL}{Imitation learning}
\acro{DAgger}{Dataset-Aggregation}
\acro{MDP}{Markov Decision Process}
\acro{VIO}{Visual-Inertial Odometry}
\acro{VSA}{Visuomotor Sampling Augmentation}
\acro{NeRF}{Neural Radiance Field}
\acro{RL}{Reinforcement Learning}
\acro{DA}{Data augmentation}
\acro{NN}{neural network}
\acro{CNN}{convolutional NN}
\acro{CoM}{Center of Mass}
\acro{RMSE}{Root Mean Squared Error}
\acro{RMS}{Root Mean Squared}
\acro{Tube-NeRF}{Tube-NeRF}
\end{acronym}

\end{document}